\documentclass[letterpaper, 10 pt, conference]{ieeeconf}  

\IEEEoverridecommandlockouts                              

\usepackage{graphicx, graphics} 
\usepackage{times} 
\usepackage{amsmath} 
\usepackage{amssymb}  

\usepackage{color}
\definecolor{mygray}{gray}{0.6}

\usepackage{epstopdf}
\usepackage{hyperref}
\usepackage{real}

\title{\LARGE \bf
	First Steps Towards Full Model Based Motion Planning and Control of Quadrupeds: A Hybrid Zero Dynamics Approach
}

\author{Wen-Loong Ma$^{1}$, Kaveh Akbari Hamed$^{2}$ and Aaron D. Ames$^{3}$
\thanks{*The work of WL. Ma and A. D. Ames are supported by NSF grants 1724464, 1544332 and 1724457.
The work of K. Akbari Hamed is supported by the NSF grants 1854898 and 1906727.}
\thanks{$^{1}$WL. Ma is with the department of mechanical engineering,
        California Institute of Technology, Pasadena, CA, 91125.
        {\tt\small wma@caltech.edu}}
\thanks{$^{2}$K. Akbari Hamed is with the faulty of mechanical engineering, 
	Virginia Polytechnic Institute and State University, Blacksburg, VA 24061
	{\tt\small kavehakbarihamed@vt.edu}}
\thanks{$^{3}$A. Ames is with the faculty of the department of Control + Dynamical Systems, California Institute of Technology, Pasadena, CA, 91125. {\tt\small ames@caltech.edu}}
}

\begin{document}
\maketitle
\thispagestyle{empty}
\pagestyle{empty}

\begin{abstract}
The hybrid zero dynamics (HZD) approach has become a powerful tool for the gait planning and control of bipedal robots. This paper aims to extend the HZD methods to address walking, ambling and trotting behaviors on a quadrupedal robot. We present a framework that systematically generates a wide range of optimal trajectories and then provably stabilizes them for the full-order, nonlinear and hybrid dynamical models of quadrupedal locomotion. The gait planning is addressed through a scalable nonlinear programming using direct collocation and HZD. The controller synthesis for the exponential stability is then achieved through the Poincar\'e sections analysis. In particular, we employ an iterative optimization algorithm involving linear and bilinear matrix inequalities (LMIs and BMIs) to design HZD-based controllers that guarantee the exponential stability of the fixed points for the Poincar\'e return map. The power of the framework is demonstrated through gait generation and HZD-based controller synthesis for an advanced quadruped robot, ---Vision 60, with 36 state variables and 12 control inputs. The numerical simulations as well as real world experiments confirm the validity of the proposed framework.
\end{abstract}


\section{INTRODUCTION}



Quadrupedal locomotion has a long and rich history of outstanding agility and dynamic stability without formal analysis, thanks to the multi-support nature of such systems. Some famous quadrupedal examples include, but are not limited to, the BigDog \cite{RAIBERT200810822}, Minitaur \cite{Avik18Vertical}, ANYmal \cite{Hutter16ANYmal} and Cheetah robot \cite{Boussema19Online}. State-of-the-art approaches for the controls and planning of quadrupeds mainly utilize model reduction to partly mitigate the computational complexity of the full-order techniques arising from nonlinearity and hybrid nature of models. For instance, massless legs, the linear inverted pendulum model \cite{LIPM_01} 
and planar motion planning are often utilized assumptions. While they do have many implementation advantages, one needs to design controllers that overcome the uncertainty induced by the difference between modeling and reality. Further, formal guarantees on stability are rarely studied in quadrupedal research.

On the other hand, the hybrid system approaches have become a powerful tool for modeling the dynamics of bipedal locomotion, in which, steady state locomotion corresponds to periodic solutions of these hybrid dynamical systems. One of the hybrid control approaches is the \textit{hybrid zero dynamics} (HZD) framework \cite{Jessy_Book}. HZD is an extension of the notion of Byrnes-Isidori zero dynamics \cite{Isidori_Book} to hybrid models of locomotion for which the resultant zero dynamics manifolds are invariant under the continuous- and discrete-time dynamics. HZD has been successful for designing gaits for bipedal locomotion and provide experiment-level controllers, 
see e.g., \cite{
Ames_RES_CLF_IEEE_TAC, 
Sreenath_Grizzle_HZD_Walking_IJRR,
Hamed_Buss_Grizzle_BMI_IJRR,
Tedrake_Robus_Limit_Cycles_CDC, 
Poulakakis_Grizzle_SLIP_IEEE_TAC, 
Johnson_Burden_Koditschek}, 
but has not yet been applied to the control of quadrupeds. The challenge in computation and controls mainly rises from the increased degrees of freedom (DOF) and richer contact scenarios of quadrupeds over bipeds. 

\begin{figure}[!t]		
	\vspace{2mm}
	\centering
	    \includegraphics[width=0.45\textwidth]{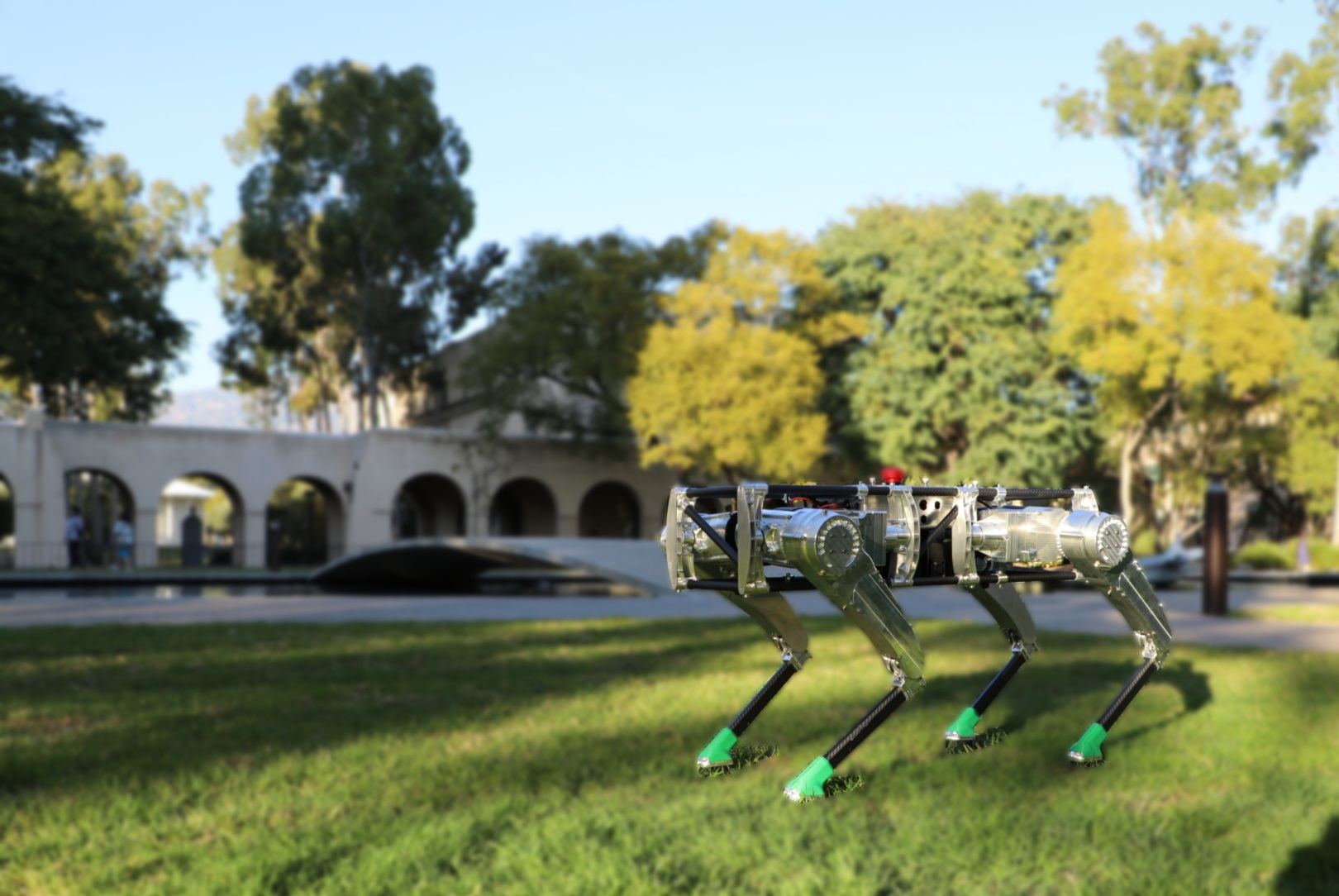}
		\caption{The Vision 60-v3.2 robot from Ghost Robotics.}
		\label{fig:robot}
	\vspace{-5mm}
\end{figure}

The main objective of this paper is to introduce an alternative approach to contemporary control schemes of quadrupeds that are mainly based on simplified models, and to make the first steps towards extending the results of HZD framework from stability critical systems such as bipeds to a more complicated but more robust system ---the quadrupedal robot. We model their dynamics as a hybrid system, optimize trajectories via a HZD optimization algorithm, analyze the dynamic stability via the Poincar\'e return map and then synthesize HZD-based controllers with iterative optimization problems including linear and bilinear matrix inequalities (LMIs and BMIs). The theoretically stable controller is validated with experiments on a quadrupedal behavior, ambling on the Vision 60 robot (\figref{fig:robot}). 
This result suggests a new approach to realize dynamically stable behaviors such as ambling, trotting and galloping in experiments.

This paper is organized as follows: Section II presents a hybrid dynamic model, based on which we design a nonlinear controller and trajectory optimization algorithm, known as the HZD optimization. Using this framework, we generate gaits including walking, ambling and trotting in simulation. Section III addresses the exponential stabilization problem as well as the HZD-based controller synthesis via BMIs. Section IV validates the ambling controllers with experiments. A comprehensive comparison is provided for the results. Section V contains conclusions and future directions. This paper presents a self-contained approach that works in both simulation and experiment, and serves as the first steps towards designing full model based controllers for quadrupedal dynamic locomotion. 
\section{Hybrid motion planning}

\begin{figure}[!t]
	\centering
	\vspace{3mm}
		\includegraphics[width=0.215\textwidth]{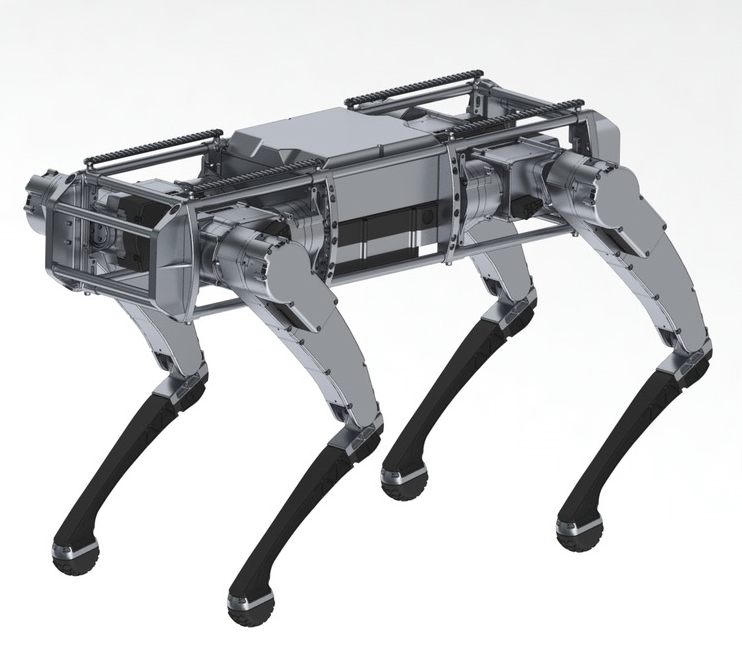}
		\includegraphics[width=0.23\textwidth]{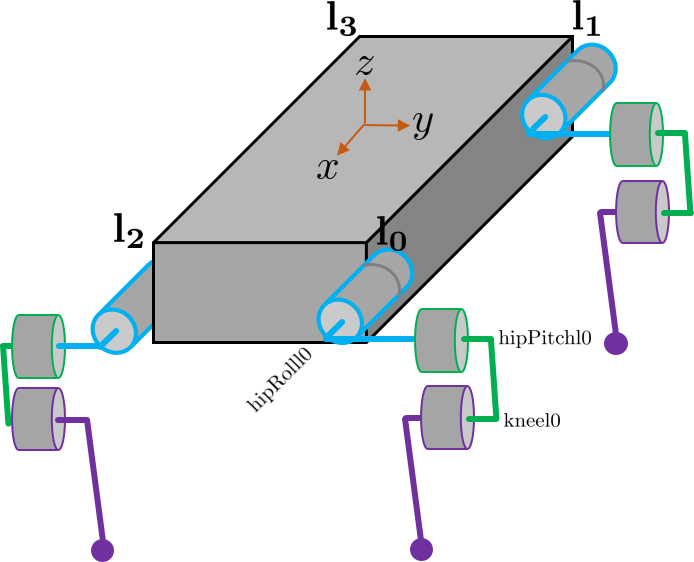}
		\caption{On the left is the rendering of the robot design, and on the right is the illustration of the configuration coordinates for the robot. The leg indices $\mathrm{l}_*$ are shown on the vertices of the \textit{body} link. Each leg has three actuated joints and equipped with a point foot.}
		\label{fig:config}
	\vspace{-5mm}
\end{figure}

We consider the nonlinear model of quadrupedal locomotion as a hybrid dynamical system, which is an alternating sequence of continuous- and discrete-time dynamics. The order of the sequence is dictated by contact events. In comparison with bipedal walking, the increased number of contact points of quadrupeds increased the complexity of the hybrid model substantially. In this section, we introduce a unified model for quadrupedal behaviors including walking, ambling and trotting, based on which we design full model based optimal controllers as well as simulation validation.


\subsection{The robot}
The robot we study in this paper is the Vision 60-v3.2 (see \figref{fig:robot}), a quadrupedal robot built by Ghost Robotics, with the total weight of $26$ kg and maximum standing height of $0.5$ m. As shown in \figref{fig:config}, we model the quadruped as a $13$-link system: a \textit{body} link and four legs, each of which has three children links ---the \textit{hip}, \textit{upper} and \textit{lower} links. The configuration variables of Vision 60 are denoted by $q\in\mathcal{Q}\subset\R^n$ where $n=18$ is the total number of degrees of freedom (DOF) without considering any contact constraints. Utilizing the floating base convention \cite{Grizzle2014Models}, we can have $q^T =( q_b^T, q_l^T )$, in which $q_b\in\R^3\times\mathrm{SO}(3)$ represents the global Cartesian position and orientation of a frame attached to the \textit{body} linkage, and the local coordinates $q_l\in\R^{12}$ denote the $12$ joint variables: hip roll, hip pitch and knee angles. These angles are denoted by $\theta_{hr_j}, \theta_{hp_j}, \theta_{k_j}$ for the $j$-th leg, all of which are actuated by Brushless (BL) DC motors. This yields the system's total DOF to be $18$ and control inputs $u\in\R^{12}$.
With different scenarios of foot contacts with the ground, we have a mixture of overactuated, fully-actuated, and underactuated domains (i.e., phases) for the dynamics. Futher, we can define the state space $\mathcal{X}=T\mathcal{Q}\subseteq \R^{2n}$ with the state vector $x^T = (q^T,\dot q^T)$, where $T\mathcal{Q}$ is the tangent bundle of the configuration space $\mathcal{Q}$. 

We now define the different modes of the quadrupedal dynamics, separated by discrete events. This fusion of continuous- and discrete-time dynamics yields a hybrid automaton model \cite{derIntro}. For more detailed definitions of hybrid modeling for bipedal robots, we refer the readers to \cite{Ames2013Humana}.

\subsection{The continuous-time domain: constrained dynamics}
Given the floating base coordinates, we can derive the unconstrained dynamics, i.e. without any contact constraints, by the Euler-Lagrange equations as:
\begin{align*}
    D(q)\, \ddot q + H(q,\dot q) &= B u
\end{align*}
where $D(q)\in\R^{n\times n}$ is the mass-inertia matrix, $ H(q,\dot q)\in\R^n$ includes the Coriolis, centrifugal and gravity terms and $B(q)\in\R^{n\times m}$ is the actuation matrix and $u\in\R^m$ is the torque applied at joints with $m=12$.

\begin{figure}[!t]
\vspace{2mm}
	\centering
		\includegraphics[width=0.45\textwidth]{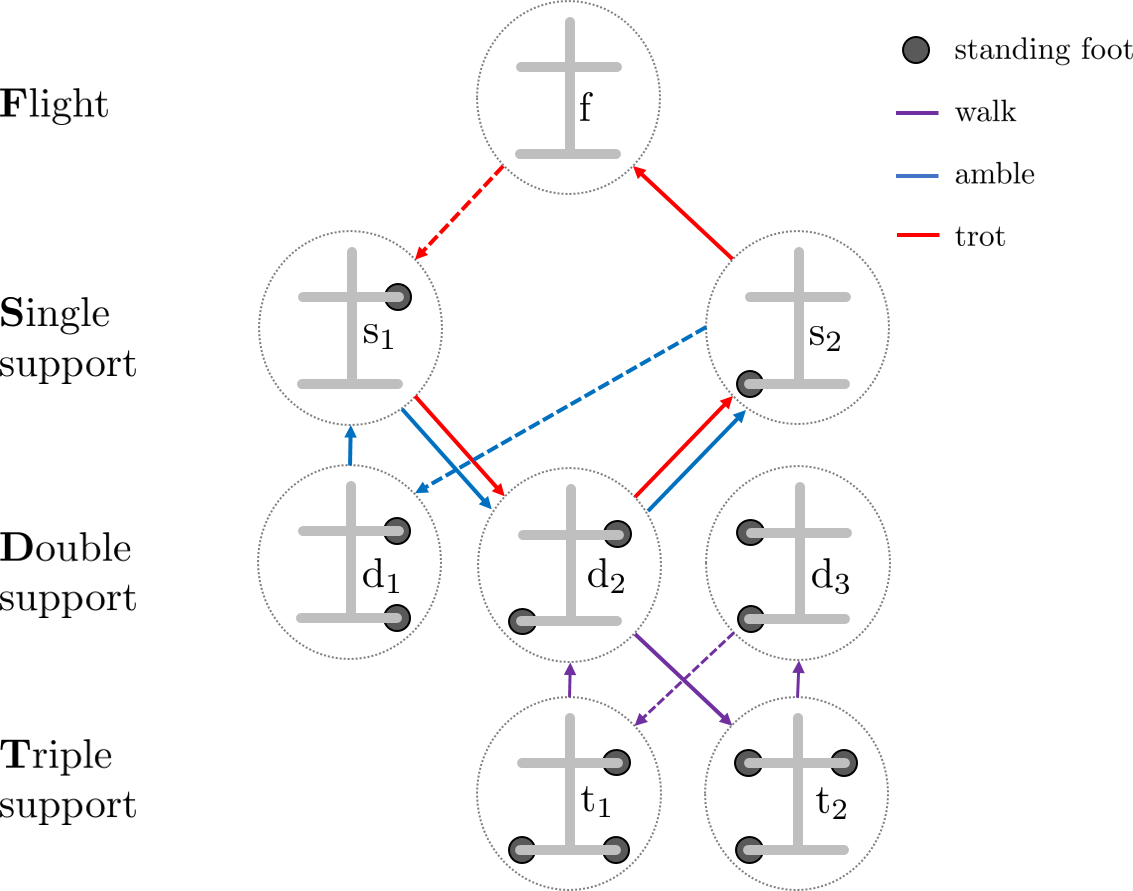}
		\caption{The cyclic directed graph for the multi-domain hybrid dynamics of walking, ambling and trotting gaits. 
		The dashed lines represent a \textit{relabeling} map \cite{Ames2013Humana} that flips the left and right legs' contact attributes.}
	\label{fig:graph}
	\vspace{-5mm}
\end{figure}

Now consider $k\in\{1,2,3,4\}$ feet standing on the ground, which means $\bar k=4-k$ feet are swinging in the air. This creates a variety of contact scenarios, hence generates different types of quadrupedal behaviors (see \figref{fig:graph}). For each contact situation, we associate a \textit{continuous domain}:

\vspace{-3mm}
\small
\begin{align}
	\label{eq:domain}
	\mathcal{D}_v := \{ (x,u) | h_v(q)=\dot h_v(q,\dot q)=0, N_v \succeq 0, h_{\mathrm{s},v}(q)\succeq 0\}
\end{align}
\normalsize
where $v\in\{1,2,3...\}$ is the domain index and $h_v(q)\in\R^k$ is the height of all standing feet with ground reaction force $N_v(x,u)\in\R^k$. The height of the other feet, referred as the swing feet, is shown by $h_{\mathrm{s}, v}(q)\in\R^{\bar k}$. 
We then have the constrained dynamics for $\mathcal{D}_v$ as: 
\begin{align}\begin{cases}
D(q)\, \ddot q + H(q,\dot q) = B_v\, u + J_{v}^T(q)\, \lambda_v \\
J_v(q)\, \ddot q  + \dot J_v(q,\dot q)\, \dot q =0
\end{cases}
\label{eq:eom-general}
\end{align}
where $J_v(q):={\partial p_v(q)} / {\partial q}$ represents the Jacobian matrix of the Cartesian position of the standing feet $p_v(q)\in\R^{3k}$, with the corresponding constraint wrench $\lambda_v\in\R^{3k}$. 
Note that the actuation matrix $B_v$ is domain dependent. This is because the double and triple support phases (\figref{fig:graph}) create closed-chain structures that induce redundancy in control and constraints. 
This is an underdetermined problem which often appears in multi-contact locomotion \cite{zhaomulti} that yields nonunique controllers for $u$. Therefore, we manually turn off the rear standing leg's hip pitch motor for double support phase and turn off the diagonal standing legs' hip pitch motors and the other standing leg's hip roll motors for triple supporting phase. This implementation in return yields underactauted dynamics for the full system. Now we can convert the constrained EOM \eqref{eq:eom-general} into a controlled ODE:
\begin{align}
\label{eq:ode}
	\dot x = f_v(x) + g_v(x) u.
\end{align}
To track a given set of time-based trajectories $\mathcal{B}_v(t)$, which will be detailed in a later section, we deployed an \textit{input output feedback linearization} controller:
\begin{align}
\label{eq:io}
	u_{\mathrm{io}}(x,t) = \mathcal{A}(x) \big(  \mathcal{L}(x,t) - 2\varepsilon y(q,t) - \varepsilon^2 \dot y(q,\dot q, t)  \big).
\end{align}
with the outputs $y(q,t) = y_a(q)-\mathcal{B}_v(t)$ and $\varepsilon>0$. In this formulation, we chose the \textit{actual outputs} $y_a(q)$ as all of the actuated joints. The notations follow directly from Eq. (28) of \cite{Ames2013Humana}. The time-based controller has been justified for improved robustness over state-based methods both in theory \cite{DURUS_timedependent} and experiments \cite{ma2017bipedal}. As a result, \eqref{eq:io} forces the system to converge to a desired gait exponentially, that is, $y_a(q)\rightarrow \mathcal{B}_v(t)$. 
The output dynamics become
\begin{align}
    \label{eq:ydyn}
	\ddot y = -2\varepsilon \dot y -\varepsilon^2 y
\end{align}
for which the origin $(y,\dot{y})=(0,0)$ is exponentially stable. 

Prior to introducing the edges and the discrete dynamics, some assumptions are necessary to construct a feasible model both for computation and experiment:
\begin{itemize}
	\item There is no ground slippage. This is partially guaranteed by enforcing a friction cone condition. However, slipping locomotion has been observed on quadrupedal animals for energy efficiency \cite{gamus2013analysis, Maslipery}.
	
	
	\item The ethology studies \cite{Muybridgeanimal} have observed a pattern of $4\times2$-domain\footnote{The term $m \times 2$ denotes a gait with $m$-domain. See \figref{fig:graph} for the directed graph of the gaits with $m$ domains. The second half of the motion is directly a left-right mirror (i.e., symmetry) of the gait. See \figref{fig:gaits} for the full motion with $m \times 2$ domains.}
	locomotion on quadrupedal animals, and the authors have investigated this modeling method in \cite{Hamed_Ma_Ames_Vision60}. But in this paper, we assume the \textit{stance leg transition domains ---one leg strikes while another leg lifts} are instantaneous and passive for walking and ambling. Hence $\mathbf{s}_1$ and $\mathbf{s}_2$ in \figref{fig:graph} become edges and we can have a $2\times 2$-domain behavior for \textit{walk} and \textit{amble}. 	
	But a $4\times 2$-domain model is still used for \textit{trot}.

\end{itemize}

\subsection{The discrete-time domain: impact and lift-off}
On the edge of $\mathcal{D}_v$ in \eqref{eq:domain}, one of the conditions reaches its bound. Thus we have two switching mechanisms:
\begin{itemize}
	\item \textit{Lift off}: a standing foot of leg ${l_*}$ lifts off from the ground, meaning 
	$N_v^{l_*}(x,u) = 0$.
	
	\item \textit{Impact}: a swing foot of leg $l_*$ impacts the ground, meaning $h_{s,v}^{l_*}(q) = 0$, $\dot h_{s,v}^{l_*}(q,\dot q) < 0$.
\end{itemize}
For \textit{lift off}, an identity map $x^+_{v+1} = x^-_{v}$ is sufficient to represent the transition from current to the next domain, where $x^-_{v}$ is the state at the end of the domain $\mathcal{D}_v$ and  $x^+_{v+1}$ is the state at the beginning of $\mathcal{D}_{v+1}$. However, the \textit{impact} shall cause a jump in the velocity terms. To describe this, we make use of a plastic impact model \cite{Grizzle2014Models},
\begin{align}
\label{eq:impact}
	\begin{bmatrix} D & -J_{v+1}^T\\ J_{v+1} &0 \end{bmatrix}
	\begin{bmatrix} \dot q^+_{v+1}\\ \Lambda \end{bmatrix}
	= 
	\begin{bmatrix} D \dot q^-_v\\ 0  \end{bmatrix}
\end{align}
by using the conservation of momentum while satisfying the next domain's holonomic constraints.

\subsection{HZD optimization}
\begin{figure*}[!t]
\vspace{2mm}
	\centering
		\includegraphics[width=0.92\textwidth]{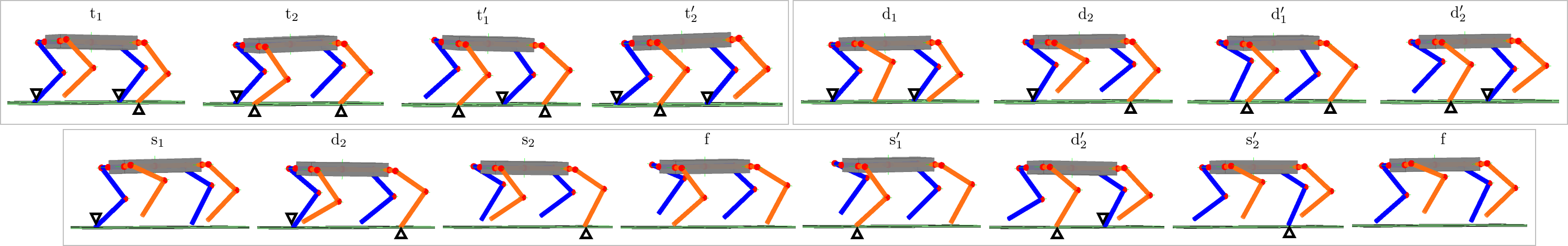}
		\vspace{-1mm}
		\caption{On the top left is the snapshot for a $2\times 2$-domain walking gait; on the top right is a $2\times 2$-domain ambling gait; and on the bottom is a $4\times 2$-domain trotting gait. The symbol $\square^\prime$ is a left-right mirror of the contact attributes for domain $\square$. 
		The orange colored legs are the left side legs whose contact points are highlighted by triangles; the blue colored legs are the right side legs with contact points highlighted by inverted triangles.}
		\label{fig:gaits}
\end{figure*}

\begin{figure*}[t!]
	\centering
		\includegraphics[width=0.3\textwidth]{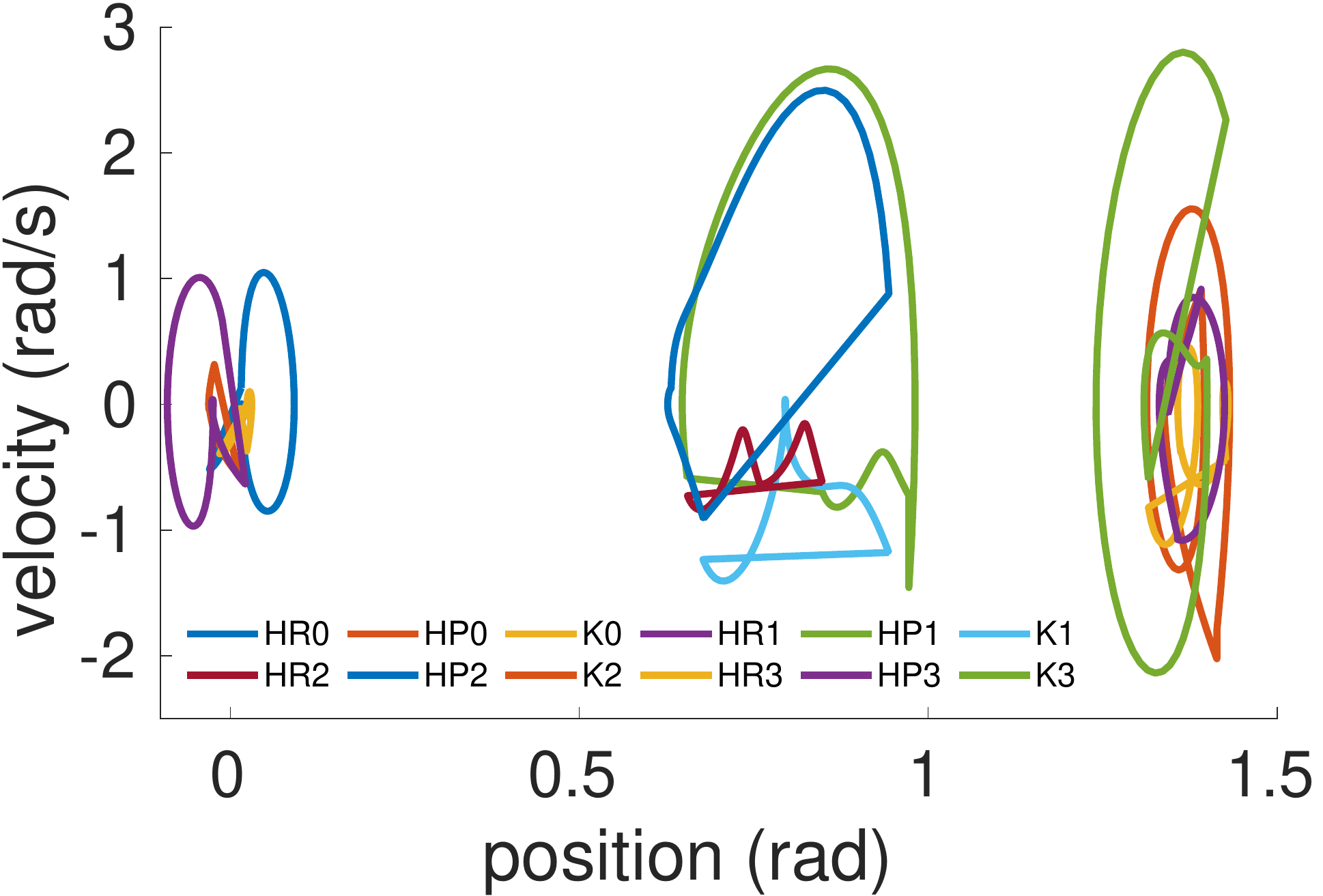}  \hspace{4mm}
		\includegraphics[width=0.3\textwidth]{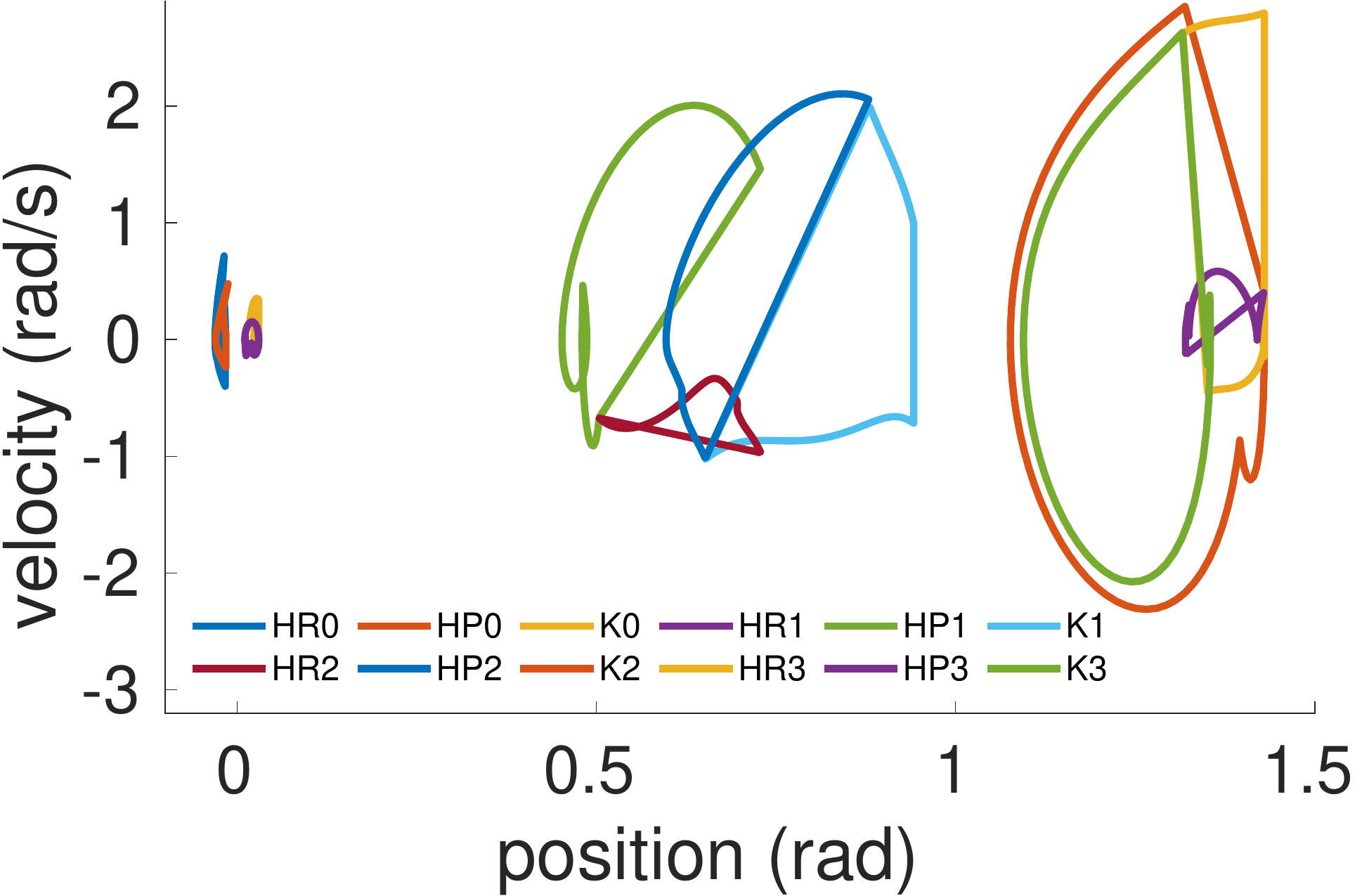} \hspace{4mm}
		\includegraphics[width=0.3\textwidth]{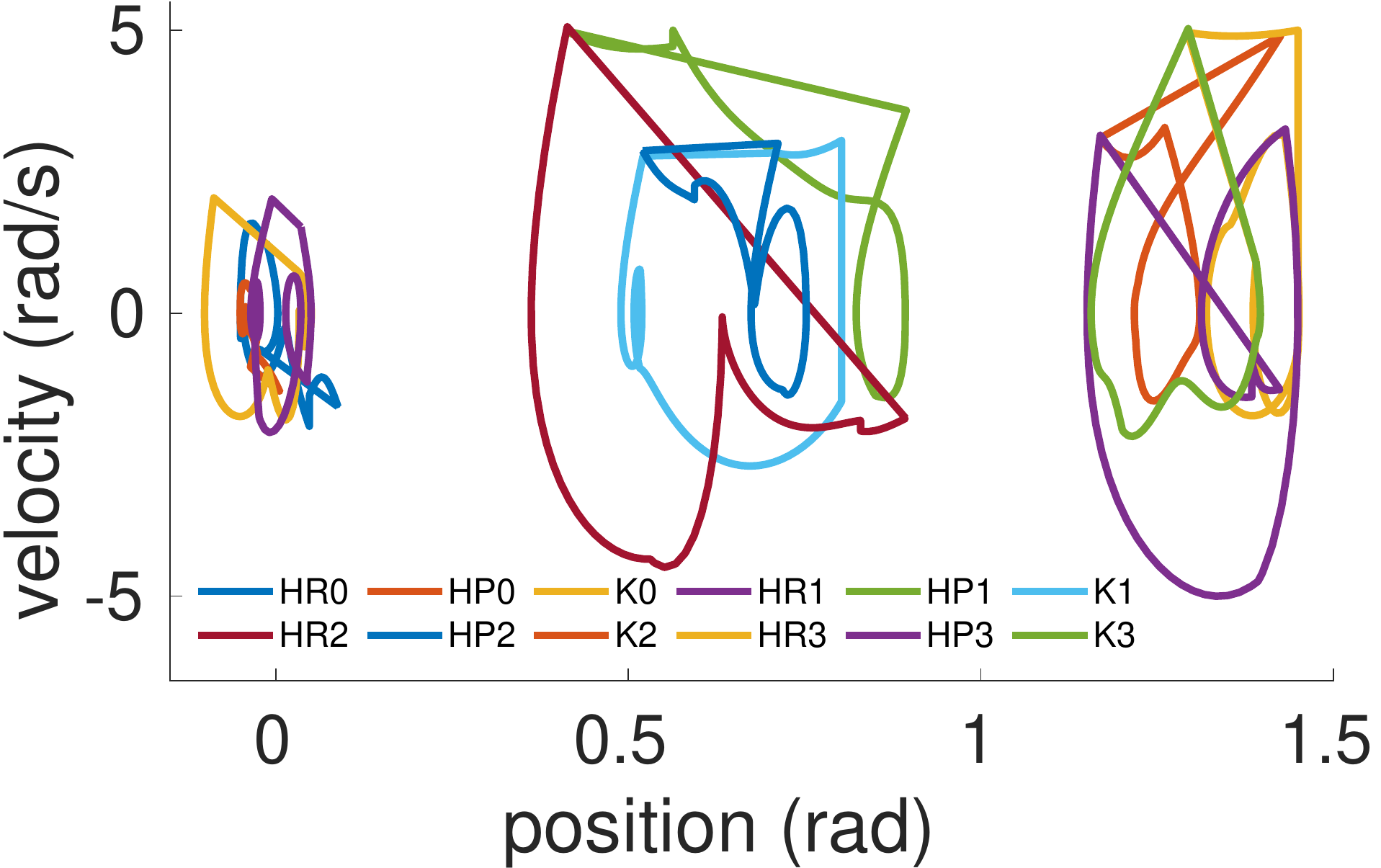}
		\caption{The periodic trajectories designed by NLP \eqref{eq:opteqs}: walk (the left), amble (the middle) and trot (the right).}
		\label{fig:gaitData}
	\vspace{-5mm}
\end{figure*}

An alternating sequence of the continuous dynamics \eqref{eq:ode} and the discrete dynamics \eqref{eq:impact} composites a hybrid control system. Since this nonlinear hybrid model has captured abundant details of the dynamics, its complexity challenges the controller design and motion planning. We hereby employ an optimization toolbox -FROST \cite{hereid2018dynamic} that parses hybrid system control problems into a nonlinear programming (NLP) based on direct collocation. 
In this approach, to generate a feasible $N$-domain motion such as walking, ambling and trotting as shown in \figref{fig:graph}, the NLP is formulated as:
\begin{align}
	\label{eq:opteqs}
	\min_{\alpha_v, x_i, \dot x_i u_i}  &\hspace{2mm}  \sum_i \norm{u_i}_2^2 \hspace{1cm}i= 1,2,...\sum_{v=1}^{N} M_v \\
	\mathrm{s.t.} 		&\hspace{3mm}  \textbf{C1}.\ \text{closed-loop dynamics} \notag \\
	&\hspace{3mm}  \textbf{C2}.\ \text{hybrid \& periodic continuity} \notag\\
	&\hspace{3mm}  \textbf{C3}.\ \text{physical feasibility} \notag
\end{align}
where $M_v$ is the number of collocation points and $\alpha_v$ is the decision variable paramterizing the desired trajectory for domain $\mathcal{D}_v$. The cost function is to minimize the torque so that experiment implementation is achievable. The constraint \textbf{C1} is from \eqref{eq:ode} and \eqref{eq:ydyn}, and the constraint \textbf{C2} is referring to the state continuity through each edge,
which could be equipped with a discrete jump in states. \textbf{C3} enforces conditions including $\norm{u_i}_{\infty} \leq 50$, $(q_i, \dot q_i)\in\mathcal{X}$, foot clearance and the friction pyramid conditions, so that the optimal solution is experimental feasible.

\textbf{Remark} Due to many types of aleatoric uncertainty in the model, not every solution of the NLP can lead to experimental success. Furthermore, some constraint setup can be ill-posed that the NLP converges poorly. Some heuristics were used to mitigate the computation:
1. Efficiently producing closed-loop controllers is the core innovation of FROST, but it also boosts the complexity of the problem. Thus using an open-loop setup (optimization without a feedback controller) to warm start a closed-loop problem is effective.
2. Tuning constraints on the acceleration (force) terms is normally more effective than tuning positions. The appearance of a behavior aligns with human intuition better, but it can often be too restrictive for the optimization to converge. 

\subsection{Optimal gaits}
Under the umbrella of the HZD framework, the sole difference among these behaviors in Fig. \ref{fig:graph} is nothing but the ordered sequence of contact events, which can be predefined by specifying the stance foot clearance as $0$ and swing foot clearance as nonzero values. Therefore, in the HZD optimization \eqref{eq:opteqs}, by changing the foot clearance constraints in \textbf{C3}, multiple quadrupedal behaviors such as \textit{walk}, \textit{amble} and \textit{trot} can be produced efficiently. We show the simulated behaviors in \figref{fig:gaits} and their phase portraits in \figref{fig:gaitData}. With some initial guesses supplied, the computation time for the presented gaits are $262.13$ s for walking, $42.69$ s for ambling and $116.05$ s for trotting on a Ubuntu $16.04$ machine with Intel Core i$7\text{-}6820$ HQ CPU @ $2.7$ GHz with $16$ GB RAM. 

\section{Exponential stabilization}

Quadrupedal dynamics are by nature contact-rich and high-dimensional. Therefore, including a Lyapunov-type stability criteria in the NLP is not computationally feasible. As a result, not every trajectory from the optimization is guaranteed to be a stable solution of the closed-loop system. In this section, we present an iterative algorithm that can postprocess the controller parameters from \eqref{eq:opteqs} to stabilize the resultant trajectories.

In the previous work \cite{Hamed_Buss_Grizzle_BMI_IJRR}, we have observed the stability of gaits in the HZD approach depends on the proper selection of the output functions to be regulated. Using the Poincar\'e sections analysis \cite{Grizzle_Asymptotically_Stable_Walking_IEEE_TAC}, our previous work \cite{Hamed_Buss_Grizzle_BMI_IJRR, Hamed_Gregg_decentralized_control_IEEE_CST} has developed a recursive algorithm, based on BMIs, to systematically design output functions for which the gaits are exponentially stable for the corresponding closed-loop dynamics. The algorithm is offline and assumes a finite-dimensional parameterization of the output functions to be determined. Then it translates the exponential stabilization problem into a recursive optimization problem that is set up based on LMIs and BMIs. The objective is to converge to a set of stabilizing parameters for which the given orbit (i.e., gait) is stable for the closed-loop system. Further, we assume that the outputs are parameterized by some controller parameters $\xi_{v}$ during the domain $v$, i.e., $y=y(q,t,\xi_{v})$. For instance, $\xi_{v}$ can parameterize the controlled variables (see \cite[Example 2]{Hamed_Buss_Grizzle_BMI_IJRR}). Then, the evolution of the hybrid system on the Poincar\'e section can be described by 
\begin{eqnarray}
x_{a}[k+1]=P_{a}\left(x_{a}[k],\xi\right),\quad k=0,1,2,\cdots,
\end{eqnarray}
where $x_{a}:=(x^\top,t)^\top$ denotes the augmented states for the nonautonomous system and $P_{a}$ represents the parameters and augmented Poincar\'e return map \cite{Parker_Book}. Here, $\xi$ denotes the controller parameters over different domains of the directed cycle. The problem of exponential stabilization consists of designing the controller parameters such that the eigenvalues of the Jacobian matrix $A(\xi):=\frac{\partial P_{a}}{\partial x_{a}}(x_{a}^\star,\xi)$  lie inside the unit circle, were $x^{\star}_{a}$ represents the corresponding fixed point for the gait. The BMI algorithm consists of three steps: 1) sensitivity analysis, 2) BMI optimization, and 3) iteration. The sensitivity analysis generates a first-order approximation for the nonlinear Poincar\'e map in terms of the controller parameters $\xi$ (see \cite[Theorem 2]{Hamed_Buss_Grizzle_BMI_IJRR}). This reduces the original exponential stabilization problem into BMIs and LMIs. More specifically, we translate the design problems of the original nonlinear system into BMIs for the first-order approximation of the Poincar\'e map \cite{Hamed_Gregg_decentralized_control_IEEE_CST}. The local solutions of the BMI optimizer are then used to update the controller parameters for the next iteration \cite{Hamed_Gregg_decentralized_control_IEEE_CST, Hamed_Grizzle_BMI_IFAC_Conference}. The algorithm continues until the spectral radius of the Jacobian linearization is less then a desired value. Sufficient conditions for the convergence of the algorithm to a set of stabilizing parameters in a finite number of iterations have been presented in our preliminary work \cite{Hamed_Gregg_decentralized_control_IEEE_CST}. 
By design, the BMI optimization problems can be solved effectively with available software packages such as PENBMI \cite{Hamed_Buss_Grizzle_BMI_IJRR,Hamed_Gregg_decentralized_control_IEEE_CST,Hamed_Gregg_IEEE_TAC,PENBMI}.

\begin{figure}[!t]
\vspace{3mm}
    \centering
    \includegraphics[width=0.21\textwidth]{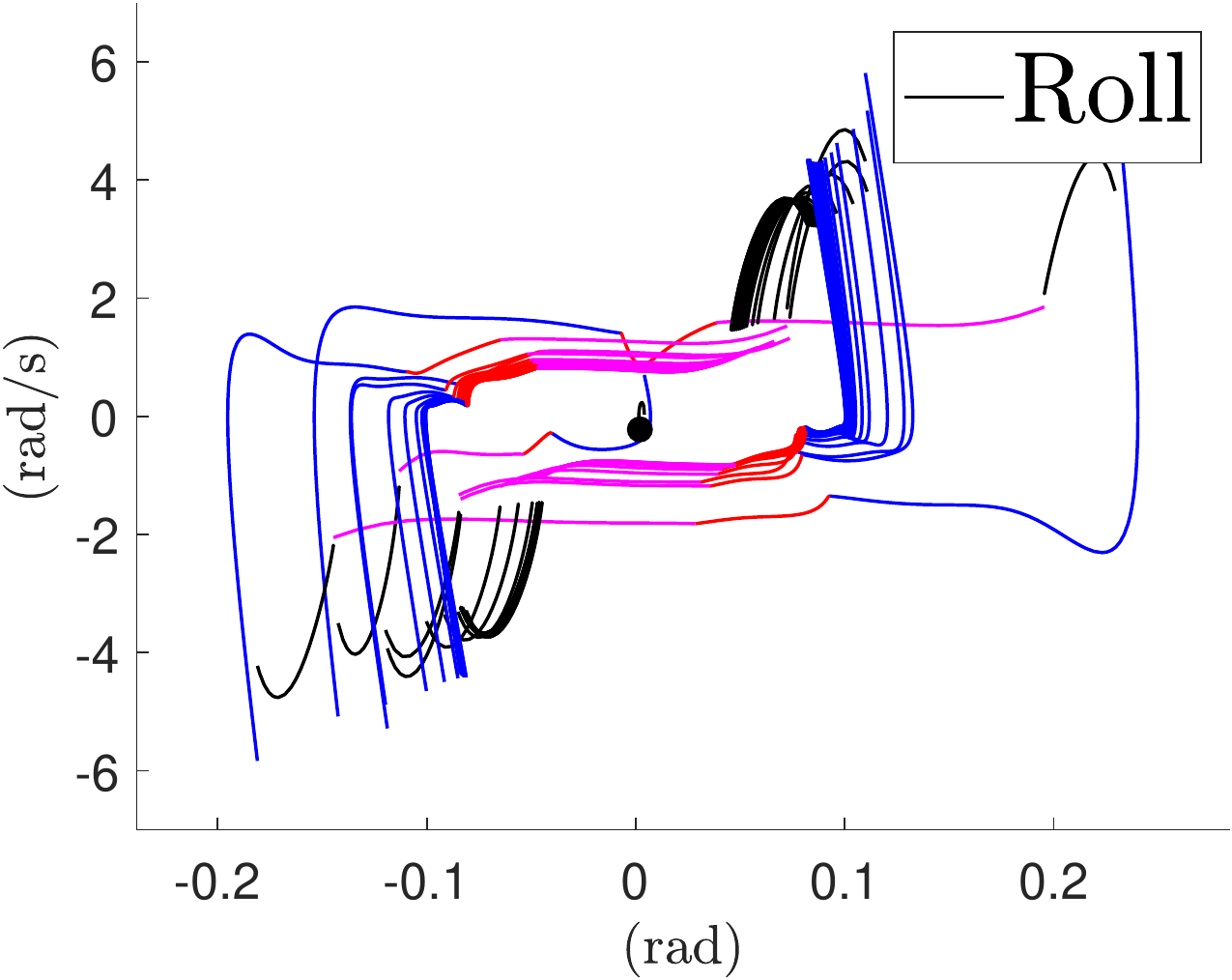}\label{roll}
    \includegraphics[width=0.21\textwidth]{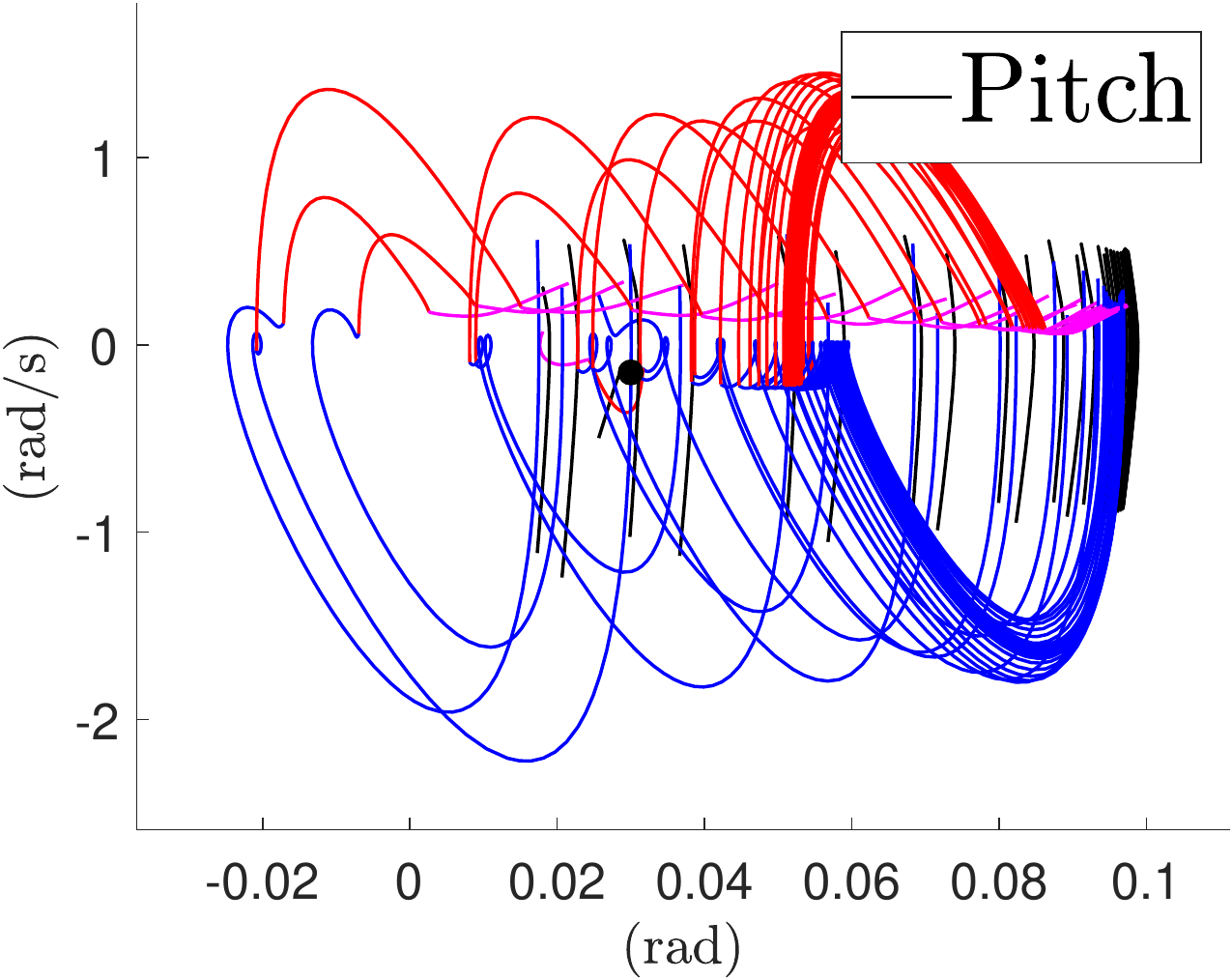}\label{pitch}
    \caption{Phase portraits during 50 consecutive steps of the 4$\times$2 domains trot gait with the BMI-optimized virtual constraint controllers.}
    \label{Phase_Portraits_BMIs}
\vspace{-3mm}
\end{figure}

We now apply this algorithm to stabilize a $4$-domain trotting gait. The dominant eigenvalues of the Jacobian matrix of the Poincar\'e map around the corresponding fixed point are $\{-1.1062,1.000, -0.4201 + 0.7278i\}$ for a \textit{trotting} gait generated by the NLP \eqref{eq:opteqs}. To stabilize it, the iterative BMI algorithm modifies the output functions to be regulated. Here, the BMI algorithm looks for the controller parameters $\xi$ that parameterize controlled variables as $y_{a}(q,\xi_{v})=H_{0}(\xi_{v})\,q$, with $H_{0}(\xi_{v})$ the output matrix that is parameterized by $\xi_{v}$. Starting from the nominal controlled variables, the BMI algorithm successfully converges to a stabilizing set of output parameters after $6$ iterations for which the dominant eigenvalues of the Poincar\'e map become $\{0.8877,-0.1507 \pm 0.8661i,-0.8830 \pm 0.0994i\}$, which locally and exponentially stabilizes the trotting gait. From the phase portraits for the closed-loop hybrid system using the BMI-stabilized controller, shown in \figref{Phase_Portraits_BMIs}, convergence to the desired orbit is clear. We remark that each iteration of the BMI optimization takes approximately $30$ minutes on a Windows machine with an Intel Xeon Gold $6130$ CPU @ $2.10$ GHz and $64$ GB RAM.

\section{Experiments}

\begin{figure}[!t]
\vspace{3mm}
	\begin{center}
		\includegraphics[width=0.155\textwidth]{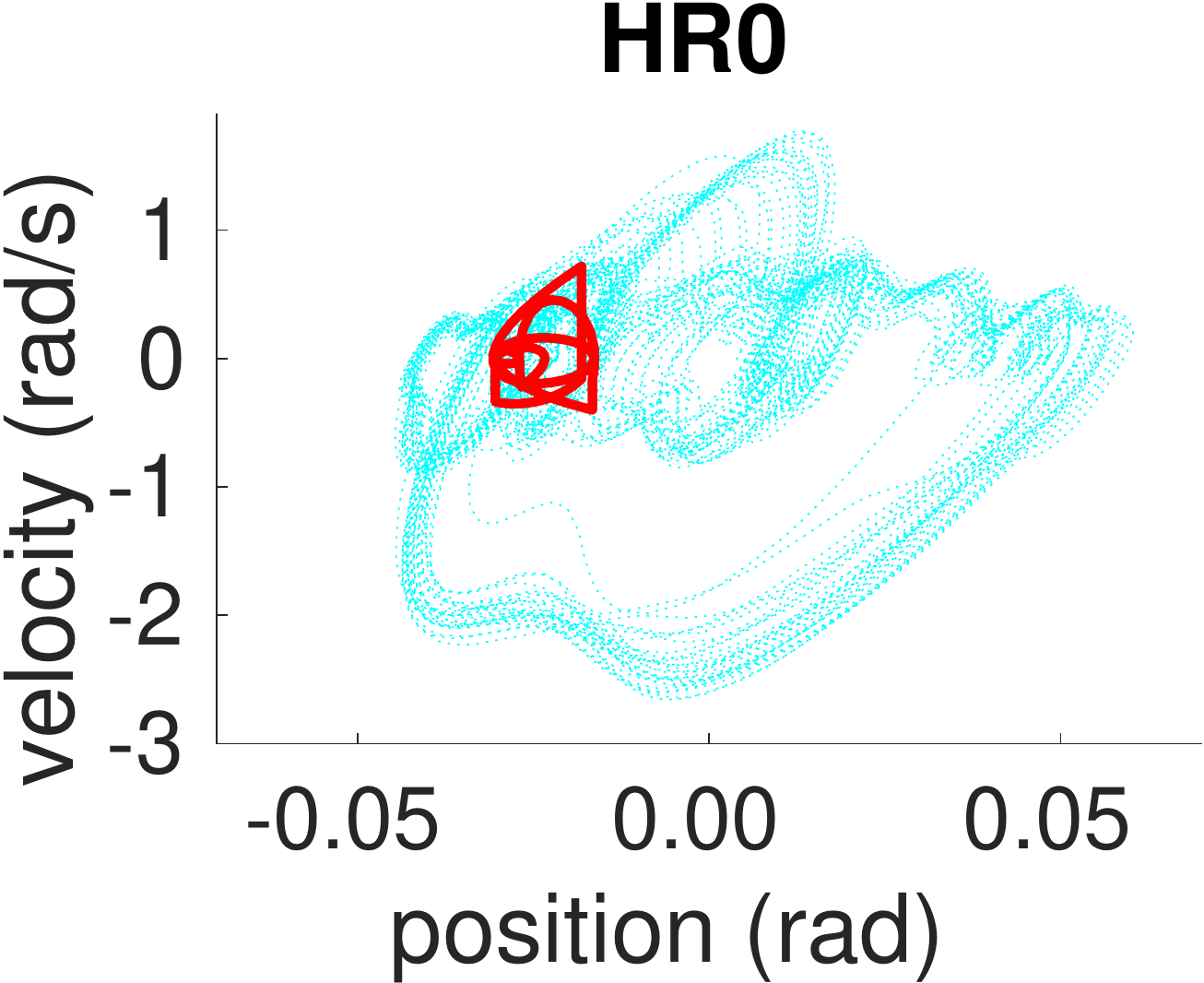}
		\includegraphics[width=0.155\textwidth]{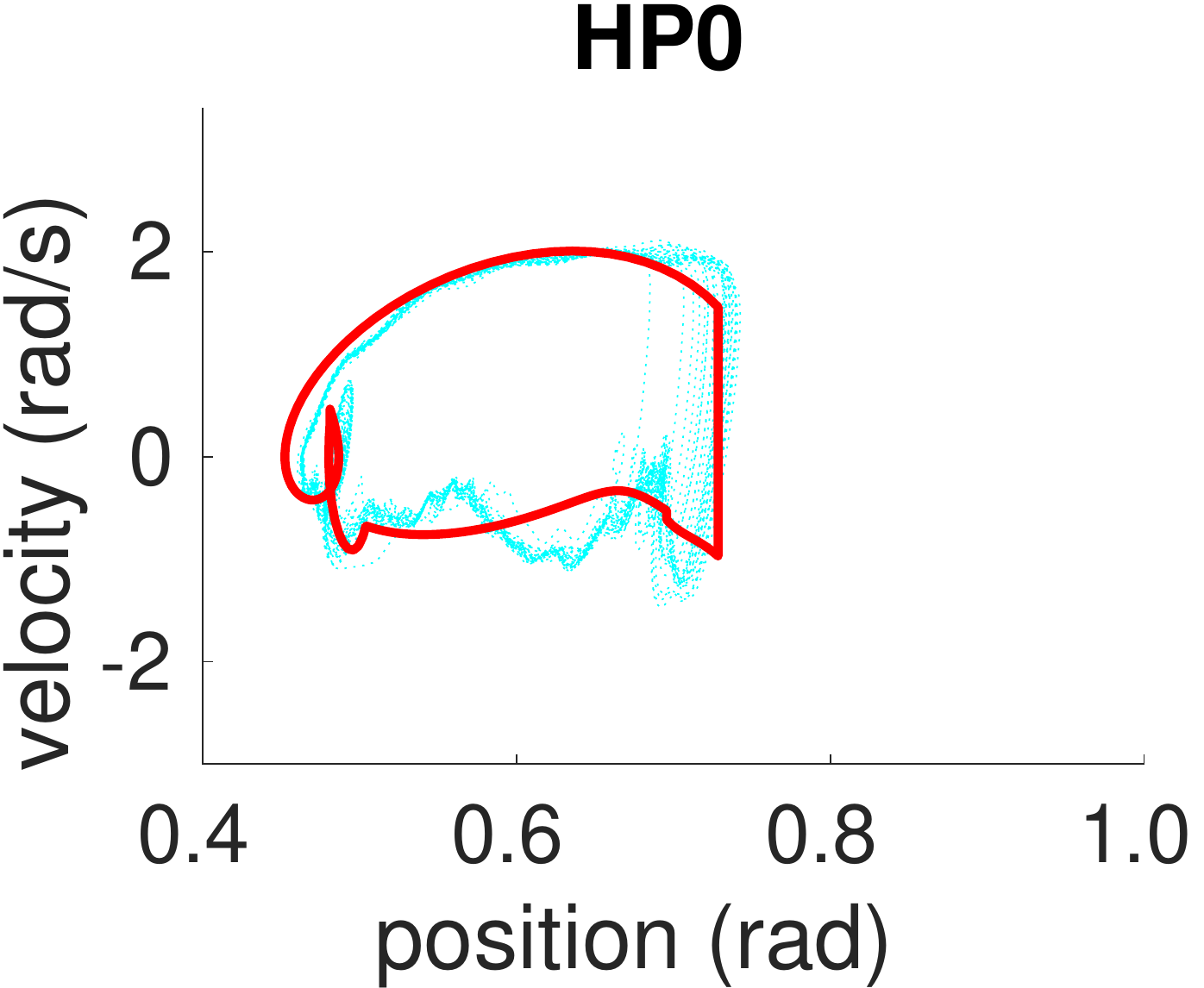}
		\includegraphics[width=0.1525\textwidth]{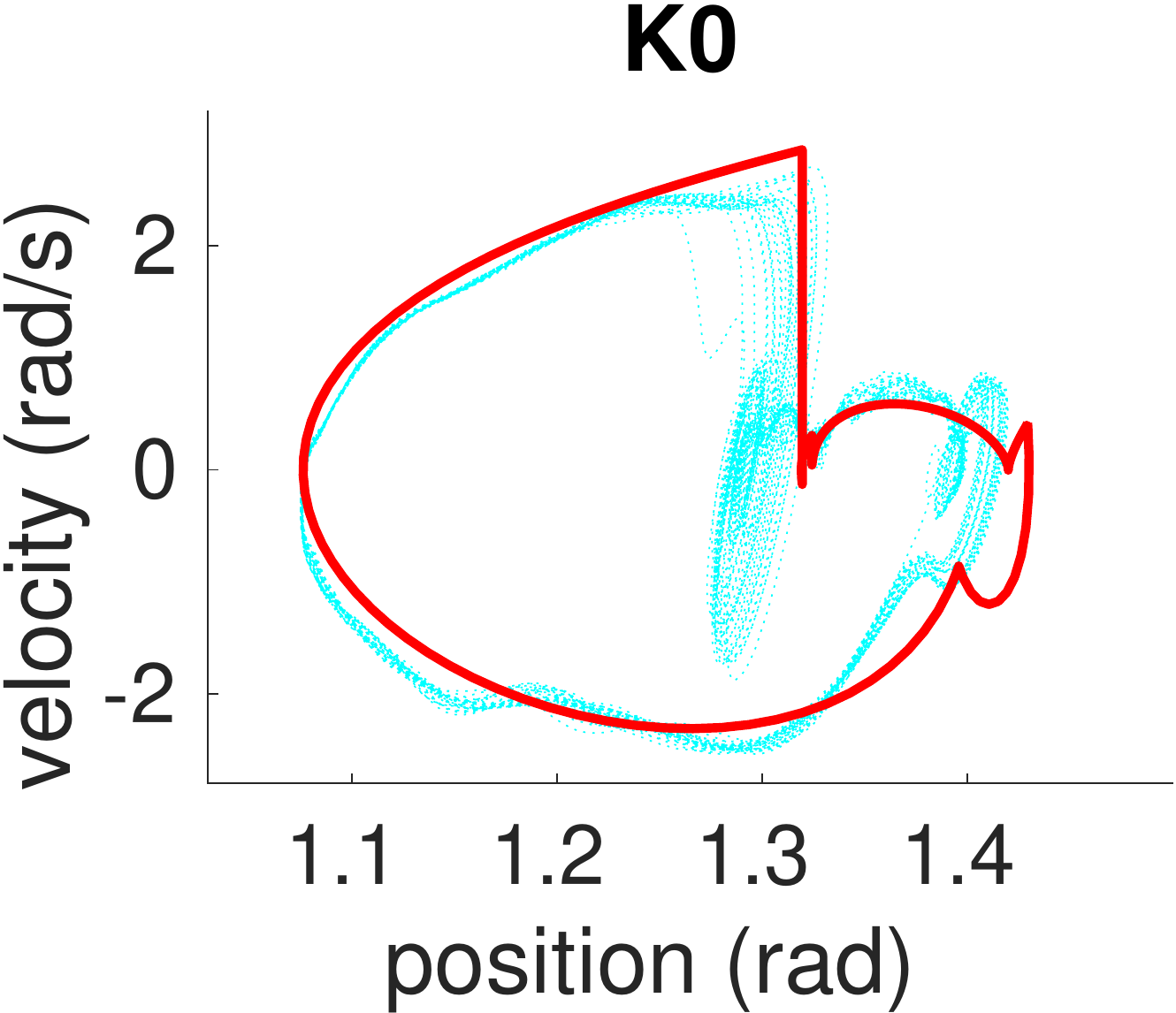}\\
		\vspace{2mm}
		\includegraphics[width=0.155\textwidth]{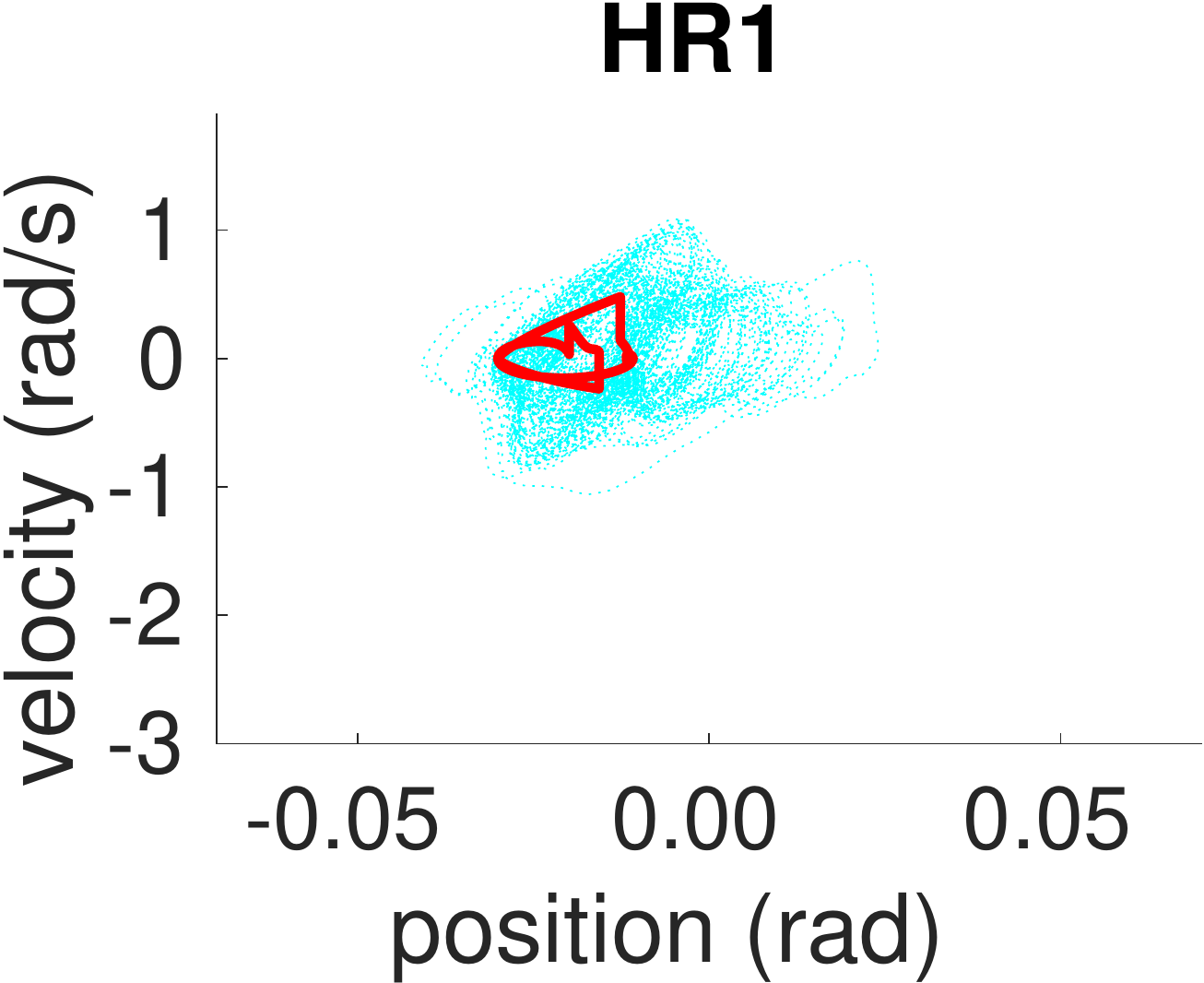}
		\includegraphics[width=0.155\textwidth]{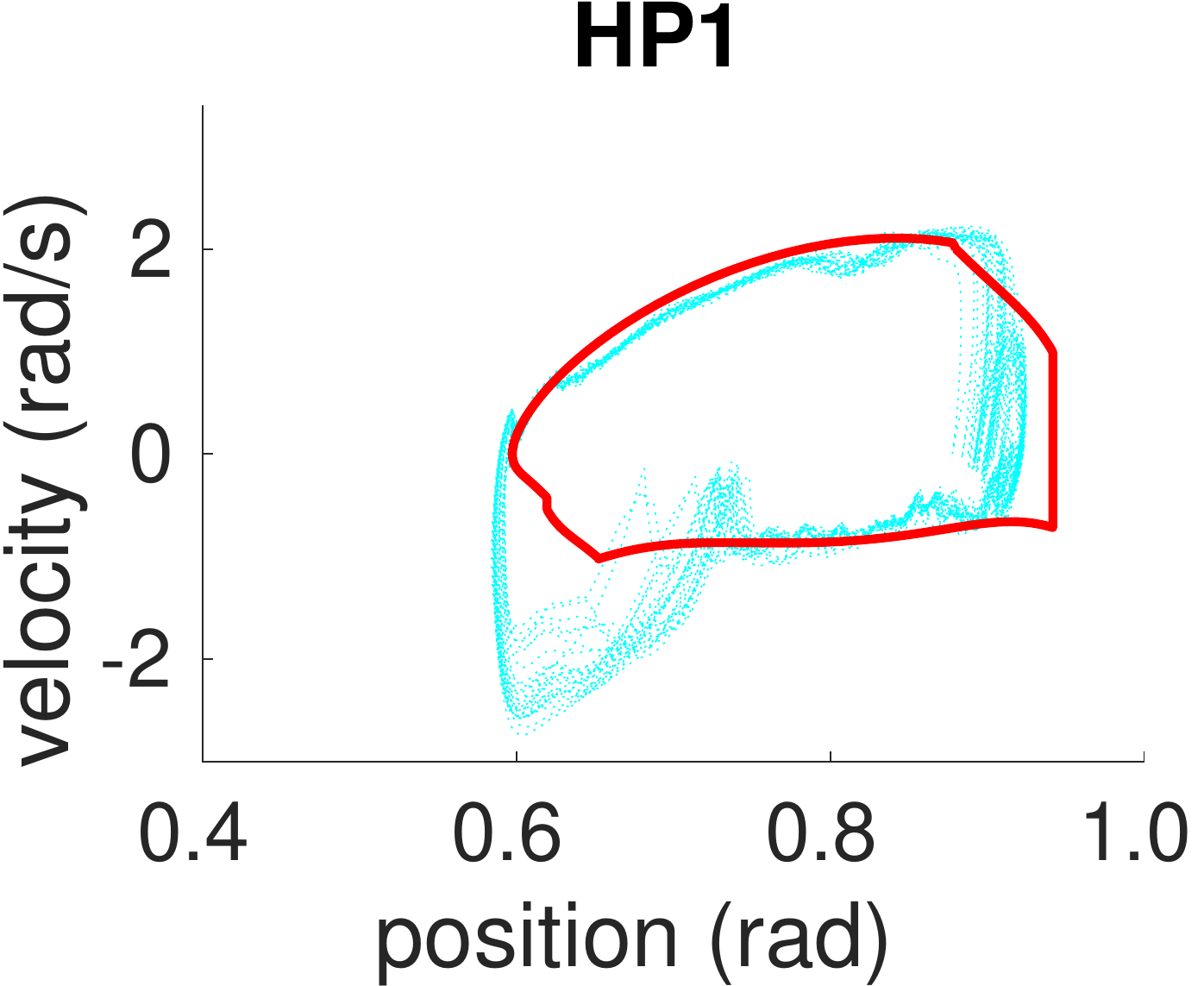}
		\includegraphics[width=0.155\textwidth]{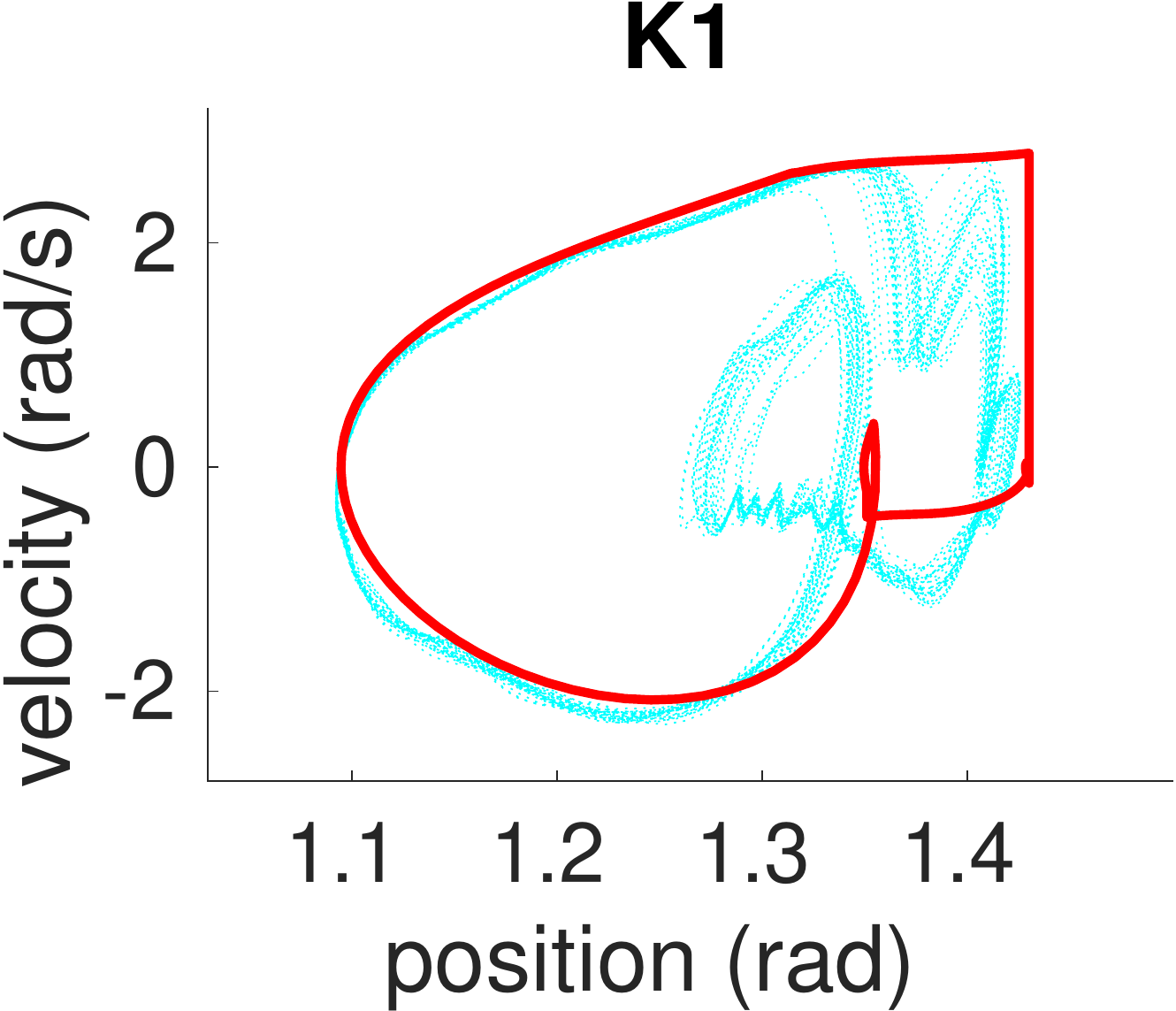}\\
		\vspace{2mm}
		\includegraphics[width=0.15\textwidth]{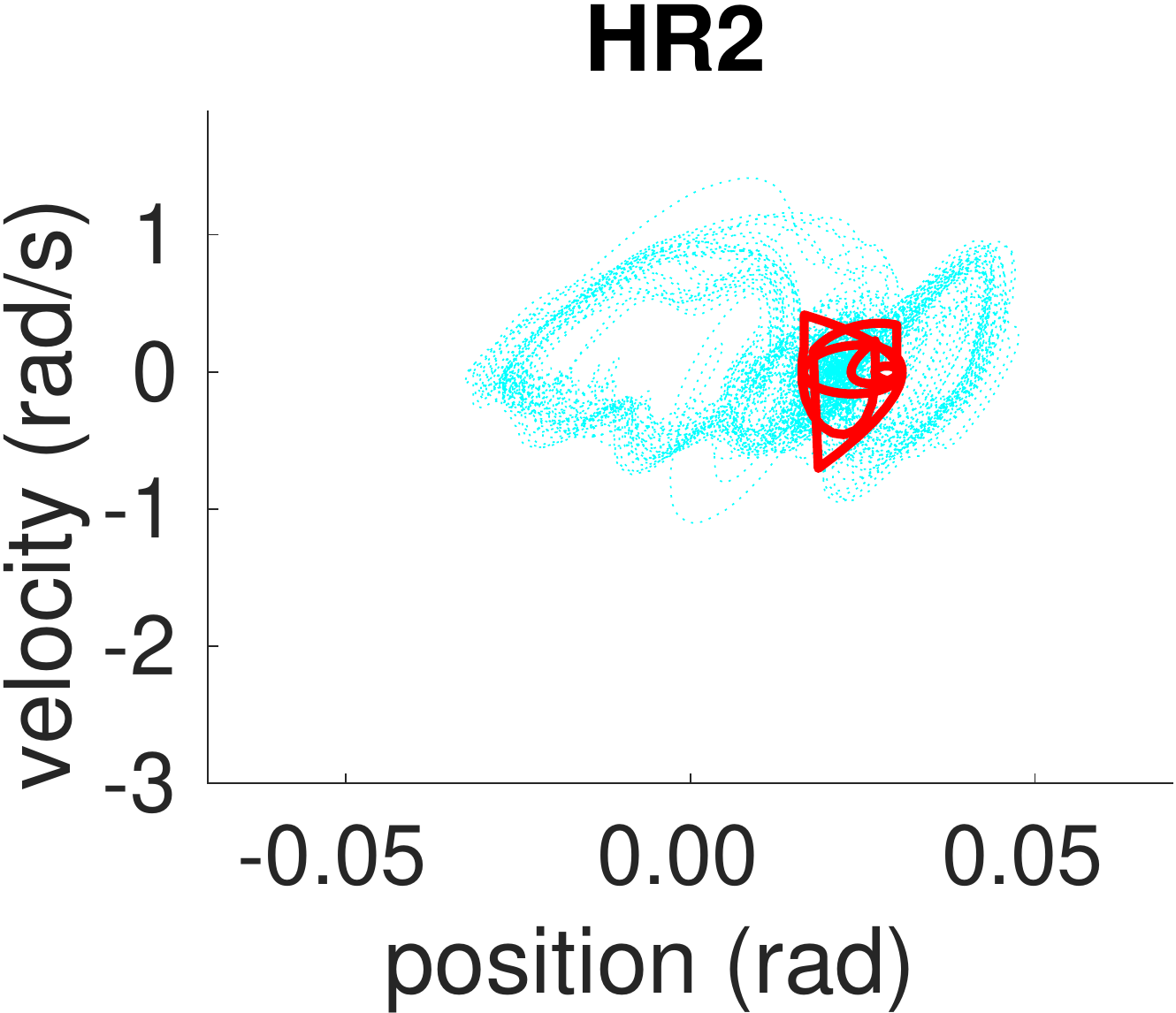}
		\includegraphics[width=0.16\textwidth]{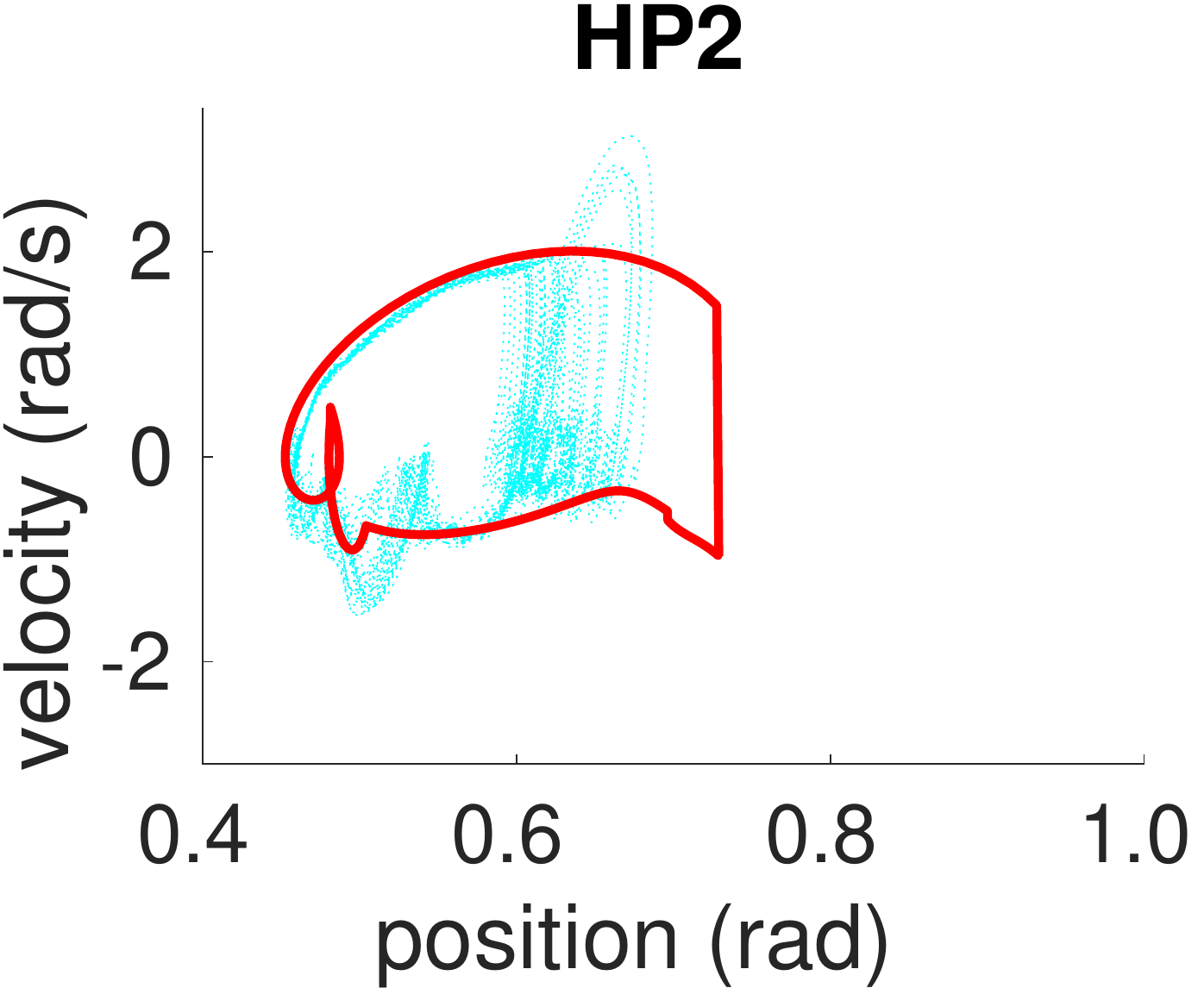}
		\includegraphics[width=0.155\textwidth]{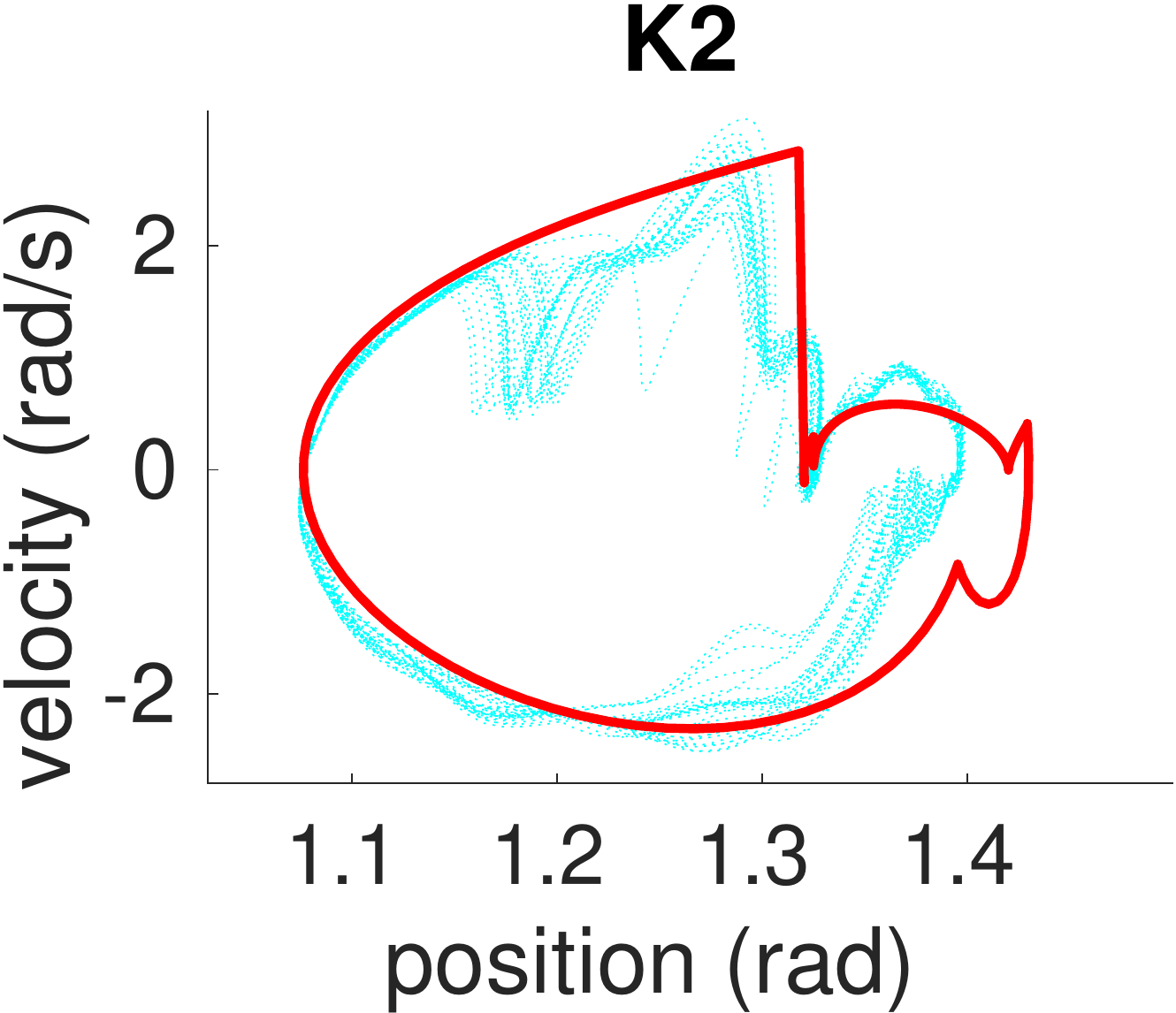}\\
		\vspace{2mm}
		\includegraphics[width=0.15\textwidth]{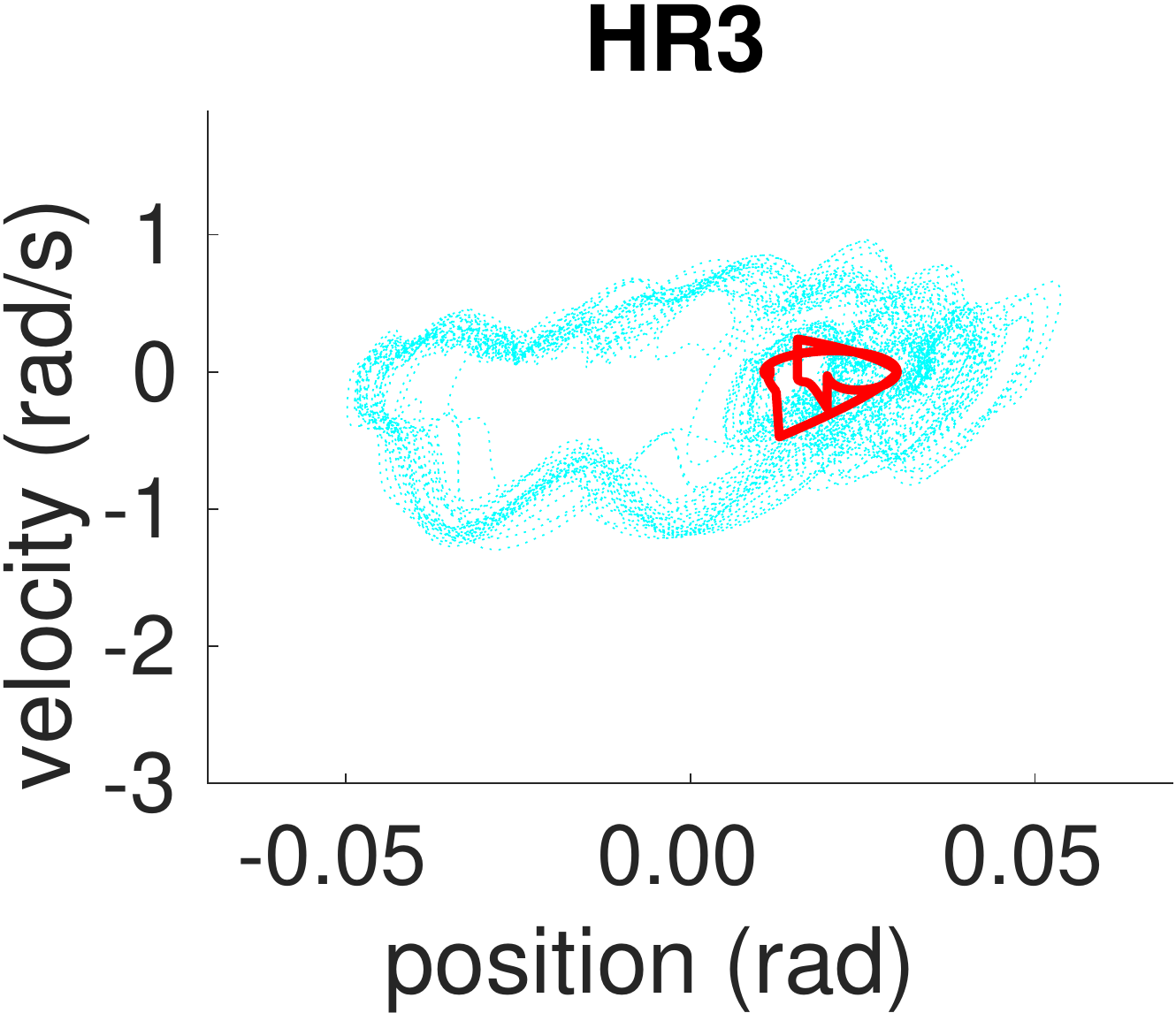}
		\includegraphics[width=0.16\textwidth]{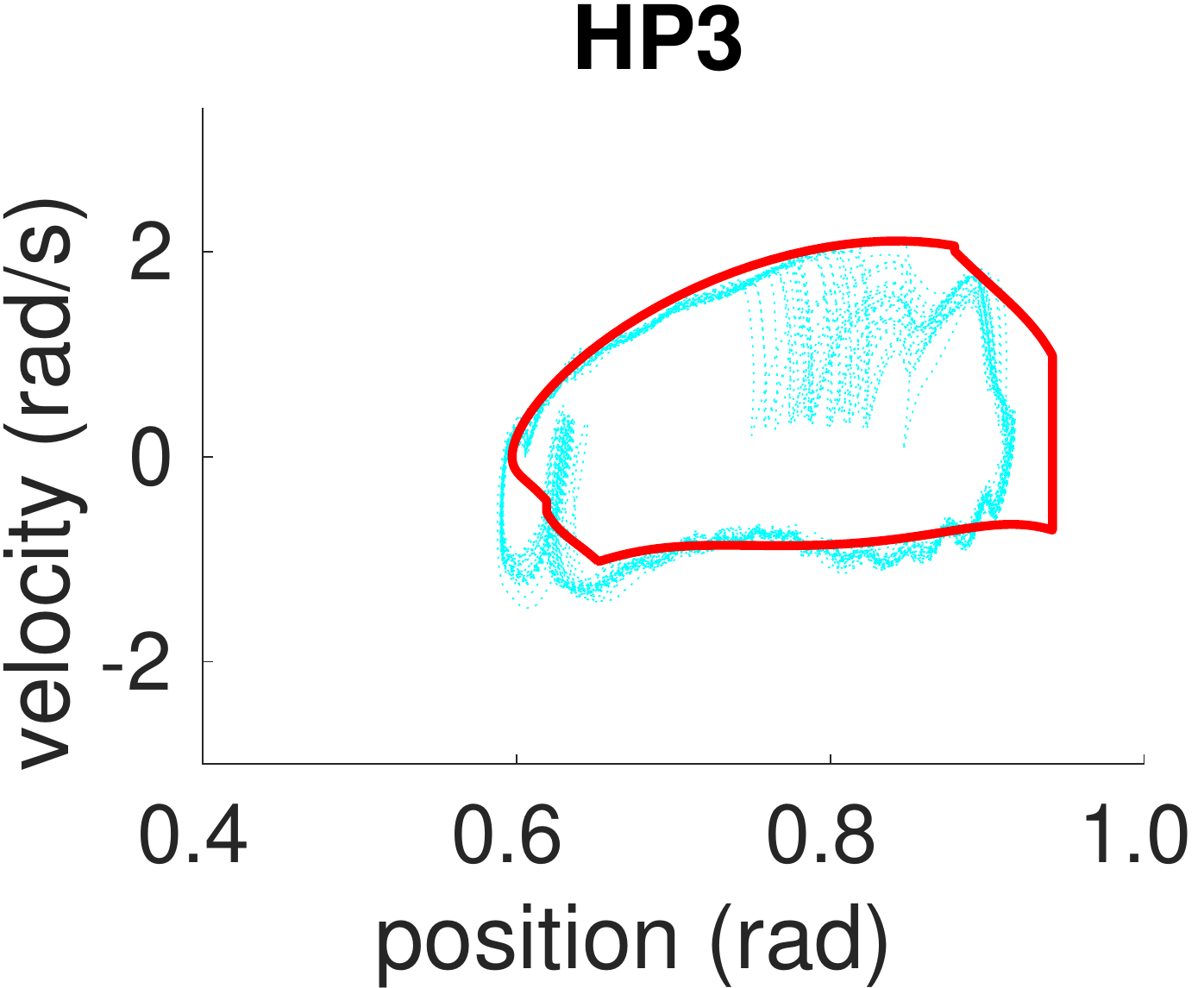}
		\includegraphics[width=0.155\textwidth]{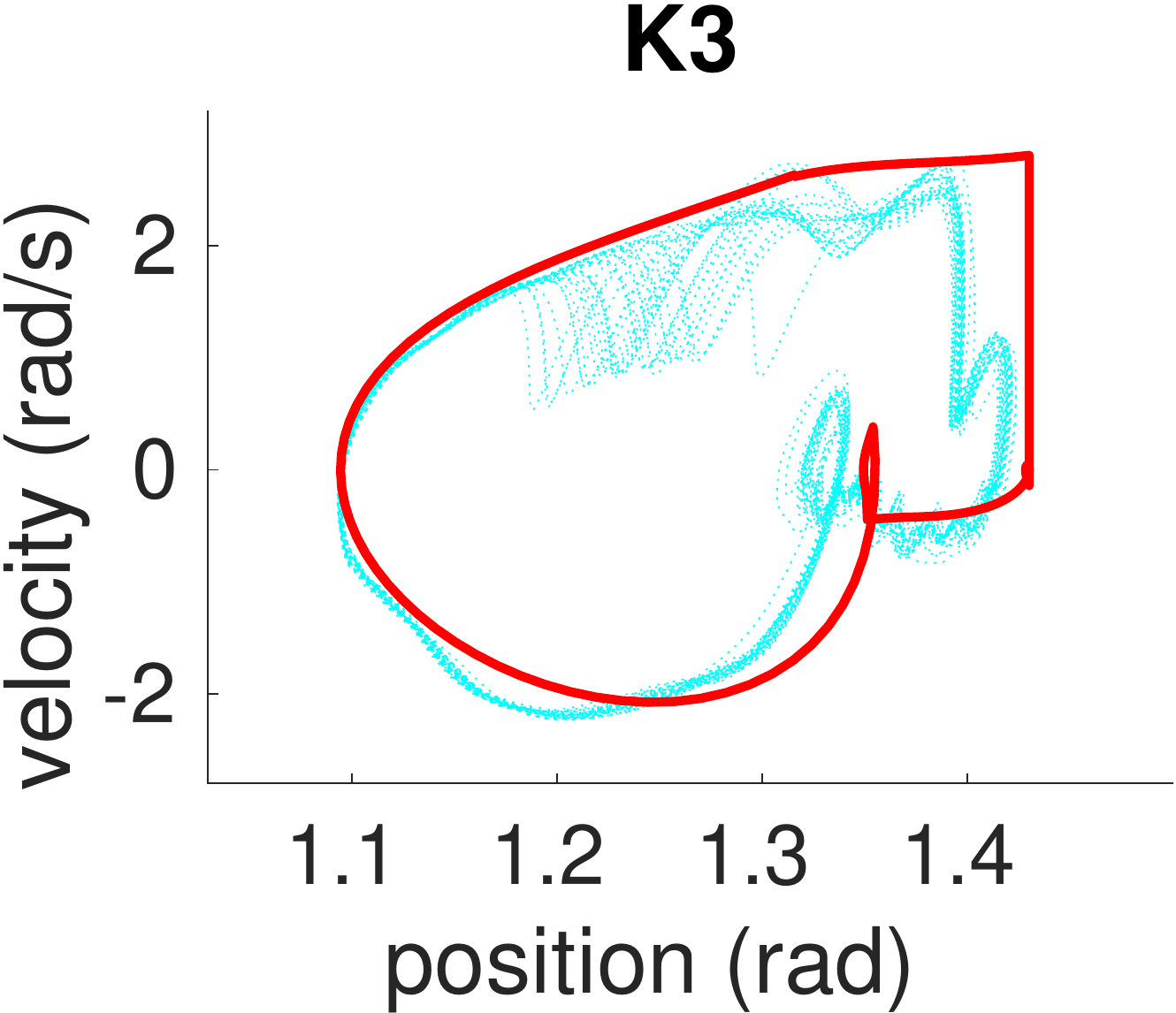}
	\caption{The designed gaits (in red) from optimization/simulation vs. the experimental data (in cyan) in the form of phase portrait for \textit{amble}. HR is short for hip roll, HP is for hip pitch and k is for knee.}
	\label{fig:walk2expsim}
	\end{center}
	\vspace{-5mm}
\end{figure}

\begin{figure*}[!t]
\vspace{3mm}
	\begin{center}
		\includegraphics[width=0.95\textwidth]{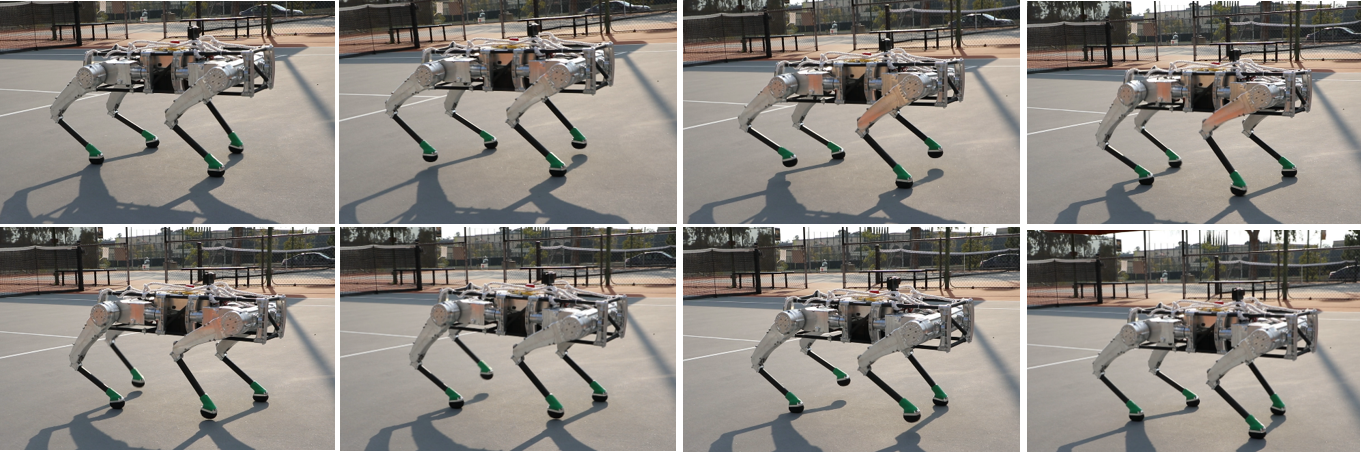}
		\caption{Snapshots of the Vision 60 ambling in an outdoor environment, showing a full step of $2\times2$ domains of the amble gait.}
		\label{fig:expTiles}
	\end{center}
	\vspace{-3mm}
\end{figure*}

With these stabilized quadrupedal dynamic gaits in simulation and optimization, we conduct an experiment with the ambling gait on Vision 60. The implemented controller is a PD approximation of the input-output linearizing controllers to track the time-based trajectories given by the optimization \eqref{eq:opteqs}. That is, for a continuous domain $\mathcal{D}_v$, we have
\begin{align}
    \label{eq:PD}
    u(q_a, \dot q_a, t) = -K_p\big(q_a - \mathcal{B}_v(t) \big) - K_d\big( \dot q_a - \mathcal{\dot{B}}_v(t) \big)
\end{align}
as the motor torque commands sent to each joint. The time-based PD implementation \eqref{eq:PD} has been shown to have exceptional robustness for bipedal locomotion \cite{DURUS_timedependent}. In addition, the domain switching method is also time-based with the event function given by the optimized trajectories. 

The result is that Vision 60 ambles stably with the desired speed of $0.3$ m/s. See \cite{amble_video} for Vision 60 ambling in an outdoor tennis court and the snapshots are shown in \figref{fig:expTiles}. We logged $20$ seconds of data and compared it with the simulated ambling controller in \figref{fig:walk2expsim}. Additionally, the average torque inputs (absolute value) are $7.73$ N$\cdot$m on the hip roll joints, $9.46$ N$\cdot$m on the hip pitch joints and $16.17$ N$\cdot$m on the knee joints. It is worth mentioning that the consistent drifting aside is expected, as there is no feedback information for the uneven terrain and that some manufacturing defects could cause asymmetric weight distribution. Some common solutions to avoid the drift is to use a joystick to manually offset the hip roll joints or \textit{Raibert-type regulators} \cite{Raibert01061984}. However, for the sake of showing a direct story from modeling and motion planning to experiments, we present the immediate implementation without any hidden layers of add-on controllers. The fact that the robot keeps ambling without falling supports the feasibility of the full model based HZD methods for quadrupedal locomotion. 

\section{Conclusion}
In this paper, we adopted the HZD framework from bipedal to quadrupedal robots. This method systematically addresses 1) the construction of multi-domain hybrid models for dynamic locomotion, 2) full-order gait planning for agile behaviors --- \textit{walk}, \textit{amble} and \textit{trot}, and 3) HZD-based control synthesis to achieve dynamic stability and robustness. We also validated one of the behaviors via experiments on a quadrupedal robot with $36$ state variables and $12$ control inputs. The result is success ambling with Vision 60. We have shown the scalability and feasibility of the HZD approaches for the controls of quadrupedal locomotion, which initiates the first steps towards realizing more dynamical behaviors. The future work includes formally defining robustness for quadrupedal locomotion, and designing optimal controllers that can allocate force through actuators for the frequently appeared overactuated scenarios.

\bibliographystyle{abbrv}
\bibliography{sec_cite}

\begin{thebibliography}{10}

\bibitem{amble_video}
Vision 60 Experiments, \url{https://youtu.be/MQCPOKCop8Q}.

\bibitem{Hamed_Buss_Grizzle_BMI_IJRR}
K.~Akbari~Hamed, B.~Buss, and J.~Grizzle.
\newblock Exponentially stabilizing continuous-time controllers for periodic
  orbits of hybrid systems: Application to bipedal locomotion with ground
  height variations.
\newblock {\em The International Journal of Robotics Research}, 35(8):977--999,
  2016.

\bibitem{Hamed_Gregg_decentralized_control_IEEE_CST}
K.~Akbari~Hamed and R.~D. Gregg.
\newblock Decentralized feedback controllers for robust stabilization of
  periodic orbits of hybrid systems: Application to bipedal walking.
\newblock {\em Control Systems Technology, IEEE Transactions on},
  25(4):1153--1167, July 2017.

\bibitem{Hamed_Gregg_IEEE_TAC}
K.~Akbari~Hamed and R.~D. Gregg.
\newblock Decentralized event-based controllers for robust stabilization of
  hybrid periodic orbits: Application to underactuated 3d bipedal walking.
\newblock {\em IEEE Transactions on Automatic Control}, pages 1--16, July 2018.

\bibitem{Hamed_Grizzle_BMI_IFAC_Conference}
K.~Akbari~Hamed and J.~Grizzle.
\newblock Iterative robust stabilization algorithm for periodic orbits of
  hybrid dynamical systems: Application to bipedal running.
\newblock In {\em The IFAC Conference on Analysis and Design of Hybrid
  Systems}, pages 161--168, Oct 2015.

\bibitem{Hamed_Ma_Ames_Vision60}
K.~Akbari~Hamed, W.-L. Ma, and A.~D. Ames.
\newblock Dynamically stable {3D} quadrupedal walking with multi-domain hybrid
  system models and virtual constraint controllers.
\newblock In {\em American Control Conference, arXiv preprint
  arXiv:1810.06697}, July 2019.

\bibitem{Ames_RES_CLF_IEEE_TAC}
A.~Ames, K.~Galloway, K.~Sreenath, and J.~Grizzle.
\newblock Rapidly exponentially stabilizing control {Lyapunov} functions and
  hybrid zero dynamics.
\newblock {\em Automatic Control, IEEE Transactions on}, April 2014.

\bibitem{Ames2013Humana}
A.~D. Ames.
\newblock Human-inspired control of bipedal walking robots.
\newblock {\em IEEE Transactions on Automatic Control}, 59(5):1115--1130, May
  2014.

\bibitem{Boussema19Online}
C.~{Boussema}, M.~J. {Powell}, G.~{Bledt}, A.~J. {Ijspeert}, P.~M. {Wensing},
  and S.~{Kim}.
\newblock Online gait transitions and disturbance recovery for legged robots
  via the feasible impulse set.
\newblock {\em IEEE Robotics and Automation Letters}, 4(2):1611--1618, April
  2019.

\bibitem{Tedrake_Robus_Limit_Cycles_CDC}
H.~Dai and R.~Tedrake.
\newblock Optimizing robust limit cycles for legged locomotion on unknown
  terrain.
\newblock In {\em Decision and Control, IEEE 51st Annual Conference on}, pages
  1207--1213, Dec 2012.

\bibitem{Avik18Vertical}
A.~De and D.~E. Koditschek.
\newblock Vertical hopper compositions for preflexive and feedback-stabilized
  quadrupedal bounding, pacing, pronking, and trotting.
\newblock {\em The International Journal of Robotics Research}, 37(7):743--778,
  2018.

\bibitem{derIntro}
S.~A. J.~v. der and J.~M. Schumacher.
\newblock {\em Introduction to Hybrid Dynamical Systems}.
\newblock Springer-Verlag, Berlin, Heidelberg, 1999.

\bibitem{gamus2013analysis}
B.~Gamus and Y.~Or.
\newblock Analysis of dynamic bipedal robot walking with stick-slip
  transitions.
\newblock In {\em ICRA}, pages 3348--3355, 2013.

\bibitem{Grizzle_Asymptotically_Stable_Walking_IEEE_TAC}
J.~Grizzle, G.~Abba, and F.~Plestan.
\newblock Asymptotically stable walking for biped robots: {A}nalysis via
  systems with impulse effects.
\newblock {\em Automatic Control, IEEE Transactions on}, 46(1):51--64, Jan
  2001.

\bibitem{Grizzle2014Models}
J.~W. Grizzle, C.~Chevallereau, R.~W. Sinnet, and A.~D. Ames.
\newblock Models, feedback control, and open problems of {3D} bipedal robotic
  walking.
\newblock {\em Automatica}, 50(8):1955 -- 1988, 2014.

\bibitem{PENBMI}
D.~Henrion, J.~Lofberg, M.~Kocvara, and M.~Stingl.
\newblock {Solving polynomial static output feedback problems with PENBMI}.
\newblock In {\em Decision and Control, and European Control Conference. 44th
  IEEE Conference on}, pages 7581--7586, Dec 2005.

\bibitem{hereid2018dynamic}
A.~Hereid, C.~M. Hubicki, E.~A. Cousineau, and A.~D. Ames.
\newblock Dynamic humanoid locomotion: A scalable formulation for {HZD} gait
  optimization.
\newblock {\em IEEE Transactions on Robotics}, 2018.

\bibitem{Hutter16ANYmal}
M.~{Hutter}, C.~{Gehring}, D.~{Jud}, A.~{Lauber}, C.~D. {Bellicoso},
  V.~{Tsounis}, J.~{Hwangbo}, K.~{Bodie}, P.~{Fankhauser}, M.~{Bloesch},
  R.~{Diethelm}, S.~{Bachmann}, A.~{Melzer}, and M.~{Hoepflinger}.
\newblock Anymal - a highly mobile and dynamic quadrupedal robot.
\newblock In {\em 2016 IEEE/RSJ International Conference on Intelligent Robots
  and Systems (IROS)}, Oct 2016.

\bibitem{Isidori_Book}
A.~Isidori.
\newblock {\em Nonlinear Control Systems}.
\newblock Springer; 3rd edition, 1995.

\bibitem{Johnson_Burden_Koditschek}
A.~M. Johnson, S.~A. Burden, and D.~E. Koditschek.
\newblock A hybrid systems model for simple manipulation and self-manipulation
  systems.
\newblock {\em The International Journal of Robotics Research},
  35(11):1354--1392, 2016.

\bibitem{LIPM_01}
S.~Kajita, K.~Tani, and A.~Kobayashi.
\newblock Dynamic walk control of a biped robot along the potential energy
  conserving orbit.
\newblock In {\em IEEE International Workshop on Intelligent Robots and
  Systems, Towards a New Frontier of Applications}, pages 789--794 vol.2, July
  1990.

\bibitem{DURUS_timedependent}
S.~Kolathaya, A.~Hereid, and A.~D. Ames.
\newblock {Time dependent control Lyapunov functions and hybrid zero dynamics
  for stable robotic locomotion}.
\newblock In {\em 2016 American Control Conference (ACC)}, pages 3916--3921,
  July 2016.

\bibitem{ma2017bipedal}
W.-L. Ma, S.~Kolathaya, E.~R. Ambrose, C.~M. Hubicki, and A.~D. Ames.
\newblock Bipedal robotic running with durus-2d: Bridging the gap between
  theory and experiment.
\newblock In {\em Proceedings of the 20th International Conference on Hybrid
  Systems: Computation and Control}, HSCC '17, pages 265--274, New York, NY,
  USA, 2017. ACM.

\bibitem{Maslipery}
W.-L. Ma, Y.~Or, and A.~D. Ames.
\newblock Dynamic walking on slippery surfaces: Demonstrating stable bipedal
  gaits with planned ground slippage.
\newblock In {\em Robotics and Automation, IEEE International Conference on},
  May 20019.

\bibitem{Muybridgeanimal}
E.~Muybridge.
\newblock {\em Animals in Motion}.
\newblock London:Chapman and Hall, LD.

\bibitem{Parker_Book}
T.~Parker and L.~Chua.
\newblock {\em Practical Numerical Algorithms for Chaotic Systems}.
\newblock Springer, 1989.

\bibitem{Poulakakis_Grizzle_SLIP_IEEE_TAC}
I.~Poulakakis and J.~Grizzle.
\newblock The spring loaded inverted pendulum as the hybrid zero dynamics of an
  asymmetric hopper.
\newblock {\em Automatic Control, IEEE Transactions on}, 54(8):1779--1793, Aug
  2009.

\bibitem{RAIBERT200810822}
M.~Raibert, K.~Blankespoor, G.~Nelson, and R.~Playter.
\newblock Bigdog, the rough-terrain quadruped robot.
\newblock {\em IFAC Proceedings Volumes}, 41(2):10822 -- 10825, 2008.
\newblock 17th IFAC World Congress.

\bibitem{Raibert01061984}
M.~H. Raibert, H.~B. Brown, and M.~Chepponis.
\newblock Experiments in balance with a 3d one-legged hopping machine.
\newblock {\em The International Journal of Robotics Research}, 3(2):75--92,
  1984.

\bibitem{Sreenath_Grizzle_HZD_Walking_IJRR}
K.~Sreenath, H.-W. Park, I.~Poulakakis, and J.~W. Grizzle.
\newblock Compliant hybrid zero dynamics controller for achieving stable,
  efficient and fast bipedal walking on {MABEL}.
\newblock {\em The International Journal of Robotics Research},
  30(9):1170--1193, Aug. 2011.

\bibitem{Jessy_Book}
E.~Westervelt, J.~Grizzle, C.~Chevallereau, J.~Choi, and B.~Morris.
\newblock {\em Feedback Control of Dynamic Bipedal Robot Locomotion}.
\newblock Taylor \& Francis/CRC, 2007.

\bibitem{zhaomulti}
H.-H. Zhao, W.-L. Ma, M.~B. Zeagler, and A.~D. Ames.
\newblock Human-inspired multi-contact locomotion with amber2.
\newblock In {\em Cyber-Physical Systems (ICCPS), 2014 ACM/IEEE International
  Conference on}, pages 199--210, April 2014.

\end{thebibliography}
\end{document}